\documentclass{article}

\usepackage{PRIMEarxiv}
\usepackage{multicol}

\usepackage[utf8]{inputenc} 
\usepackage[T1]{fontenc}    
\usepackage{hyperref}       
\usepackage{url}            
\usepackage{booktabs}       
\usepackage{amsfonts}       
\usepackage{nicefrac}       
\usepackage{microtype}      
\usepackage{lipsum}
\usepackage{fancyhdr}       
\usepackage{graphicx}       
\graphicspath{{media/}}     

\title{
Progress and Prospects in 3D Generative AI: A Technical Overview including 3D human
}

\author{
  Song Bai
    \thanks{This paper is finished during the internship at Qiniu} \\
  University of California, Riverside \\
  \texttt{song.bai@email.ucr.edu} \\
  \And
  Jie Li \\
  Qiniu \\
  \texttt{cpunion@gmail.com} \\
}

\begin{document}

\begin{twocolumn}[
\maketitle

\begin{abstract}
While AI-generated text and 2D images continue to expand its territory, 3D generation has gradually emerged as a trend that cannot be ignored. Since the year 2023 an abundant amount of research papers has emerged in the domain of 3D generation. This growth encompasses not just the creation of 3D objects, but also the rapid development of 3D character and motion generation. Several key factors contribute to this progress. The enhanced fidelity in stable diffusion, coupled with control methods that ensure multi-view consistency, and realistic human models like SMPL-X, contribute synergistically to the production of 3D models with remarkable consistency and near-realistic appearances. The advancements in neural network-based 3D storing and rendering models, such as Neural Radiance Fields (NeRF) and 3D Gaussian Splatting (3DGS), have accelerated the efficiency and realism of neural rendered models. Furthermore, the multimodality capabilities of large language models have enabled language inputs to transcend into human motion outputs. This paper aims to provide a comprehensive overview and summary of the relevant papers published mostly during the latter half year of 2023. It will begin by discussing the AI generated object models in 3D, followed by the generated 3D human models, and finally, the generated 3D human motions, culminating in a conclusive summary and a vision for the future.
\end{abstract}
\keywords{AIGC \and Generative AI \and Text-to-3D \and 3D Generation \and Metaverse}
\vspace{0.5cm}
]

\section{Introduction}
\vspace{-0.3cm}
Traditional 3D graphics primarily relied on methods like meshes and point clouds, involving extensive use of polygons, typically triangles, or points. However, with the advent of AI 3D rendering methods, new 3D structures’ representations are emerging, notably facilitated by AI training. A prime example is Neural Radiance Fields (NeRF) \cite{mildenhall2020nerf}, which employs a singular neural network to represent an entire 3D scene, its structure is shown in Figure \ref{fig:nerfpipe}. Another innovative approach is 3D Gaussian Splatting (3DGS) \cite{kerbl20233d} proposed in the middle of year 2023, which reconstructs 3D scenes by training Gaussian points in 3D space, shown in Figure \ref{fig:3dgspipe} and Figure \ref{fig:3dgsadapt}. Both methods allow for the obtaining of images directly from specified viewpoints.

The foundation of AI-generated 3D content, highlighted by techniques like NeRF \cite{mildenhall2020nerf} and 3DGS \cite{kerbl20233d}, differs from traditional meshes and point clouds. These traditional methods are not so suitable to the AI training pipelines. These new scene storage and rendering methods offer significant advantages: Their storage or training process typically require only input images and camera parameters, as they are end-to-end deep neural network. Where the camera parameters can also be calculated using AI methods. When required for application, the output corresponding images can be obtained by simply inputting the camera parameters. To project the 3DGS \cite{kerbl20233d} onto 2D we just need to do a projective transformation using a set of matrix multiplication which make it easier for the GPU pipeline. Most process are efficiently performed on GPUs through matrix operations.

Since the release of the review paper 'Generative AI meets 3D' \cite{li2023generative} earlier this year, an abundant amount of research papers have emerged in the domain of 3D generation. The pipeline for AI-generated 3D content is continually being optimized. High-precision 3D generation tools like MVDream \cite{shi2023mvdream} and RichDreamer \cite{qiu2023richdreamer} can achieve qualities up to 8K resolution. Example of RichDreamer can be seen in Figure \ref{fig:richdreamerexample}. On the other hand, the fastest 3D generation methods, such as Direct2.5 \cite{lu2023direct25}, can produce one models in under 10 seconds. While these high-precision models may not entirely replicate the look of the real world, the rapid progress in this field is undeniable.

\begin{figure}[hbt!]
    \centering
    \includegraphics[width=0.485\textwidth]{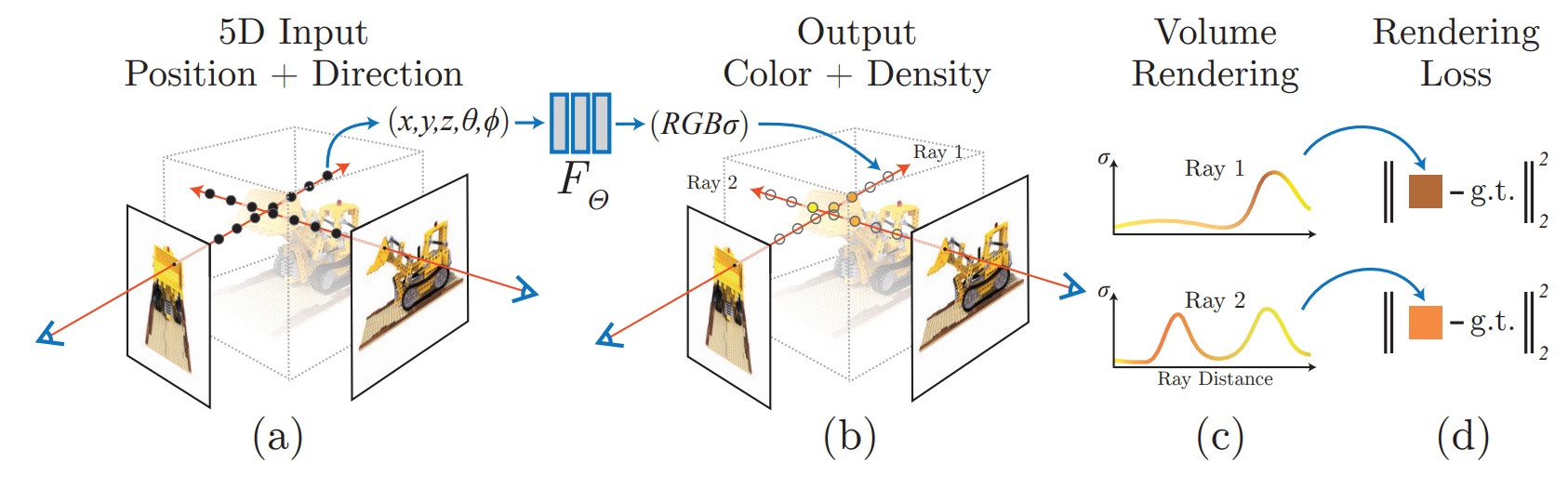}
    \caption{Structure of NeRF, picture obtained from  \cite{mildenhall2020nerf}}
    \label{fig:nerfpipe}
\end{figure}

\begin{figure}
    \centering
    \includegraphics[width=0.485\textwidth]{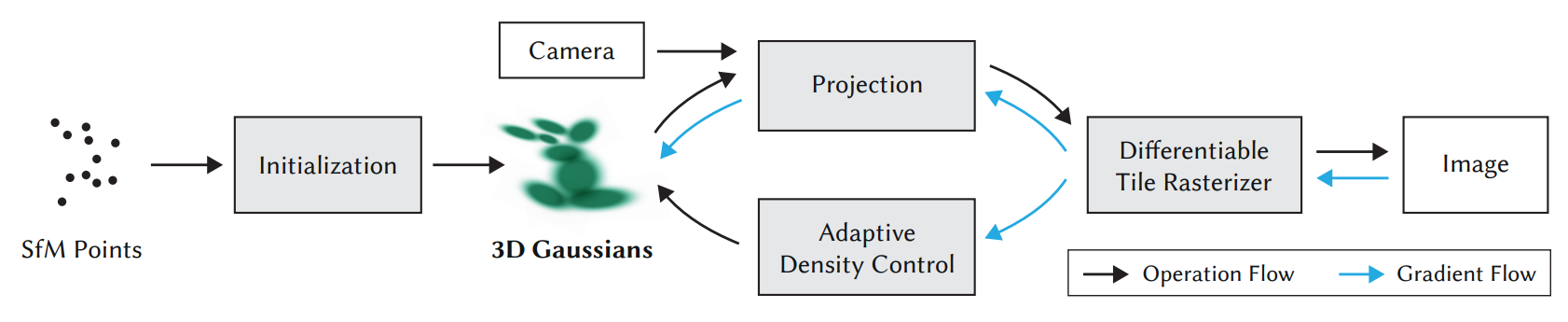}
    \caption{Structure of 3D Gaussian Splatting, picture obtained from \cite{kerbl20233d}}
    \label{fig:3dgspipe}
\end{figure}

\begin{figure}
    \centering
    \includegraphics[width=0.485\textwidth]{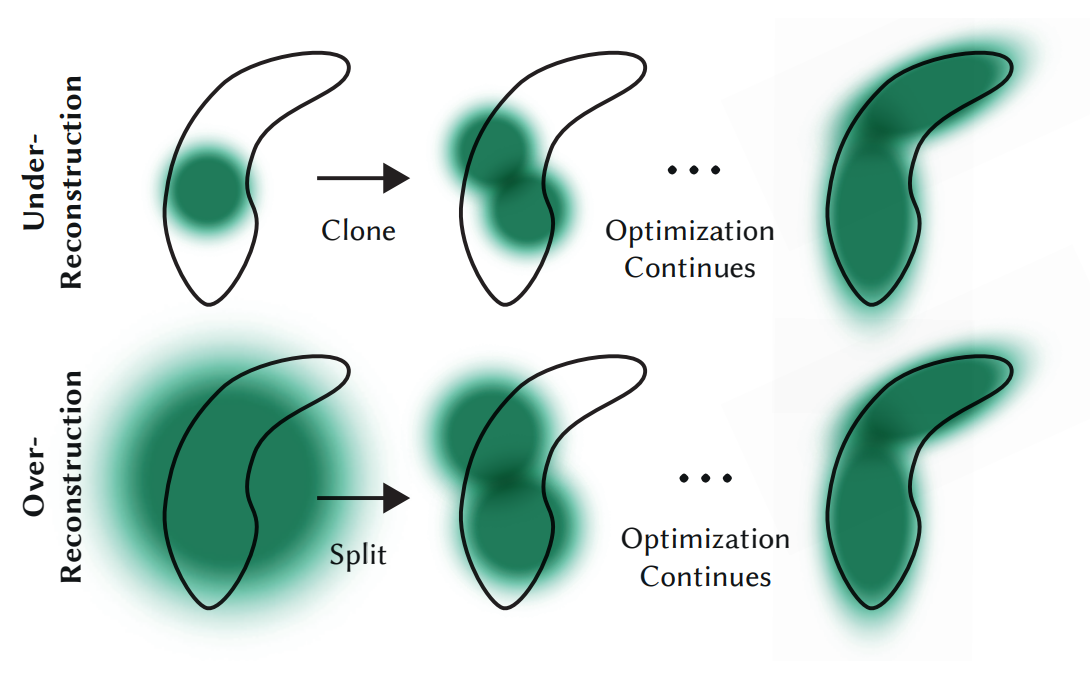}
    \caption{Adaptive Gaussian densification scheme of the 3DGS, picture obtained from \cite{kerbl20233d}}
    \label{fig:3dgsadapt}
\end{figure}

The majority of AI generated 3D modeling techniques currently utilize 2D image generation diffusion models, primarily due to the variance in 2D image dataset sizes. Notably, the ObjaverseXL \cite{deitke2023objaversexl} published at middle of year 2023, the most comprehensive 3D object dataset to date, comprises over 10.2 million objects. This is still significantly less when compared to the DiffusionDB \cite{wang2023diffusiondb} dataset, which contains 5.85 billion text-image pairs for text-to-image conversions. It's important to note that the text-3D pair dataset from last year contained only 818k entries, underscoring an anticipated rapid expansion in dataset volumes in the near future. With more data provided, the methods tends to produce greater generalizability.

Furthermore, representations of the human body have significantly advanced AI-generated 3D content, especially in human figures and movements creation. With the introduction of the Skinned Multi-Person Linear Model (SMPL) \cite{loper2015smpl} human model and subsequent research, human body models can be trained from images and outputted with high 3D realism. The body shape is mainly trained on the weight of the blend shape and joints. Due to the high fidelity of its later models like SMPL-X \cite{pavlakos2019smplx}, they are often used as a base for more refined training, making the shape control of 3D human models relatively easier than general 3D models with this prior.

Coupled with these human body representations and pre-training methods via images, technologies for synthesizing human motions through textual descriptions are also rapidly advancing. Trained large language models can understand the relationship between body movements and text, matching human motions to textual descriptions to generate time-sequenced human motions. Notable achievements, like the Story2Motion \cite{qing2023storytomotion}, GTA \cite{zhang2023globalcorrelated} project, demonstrate how a character model can walk, eat, or even dance according to linguistic descriptions.

\begin{figure*}
    \centering
    \includegraphics[width=1\textwidth]{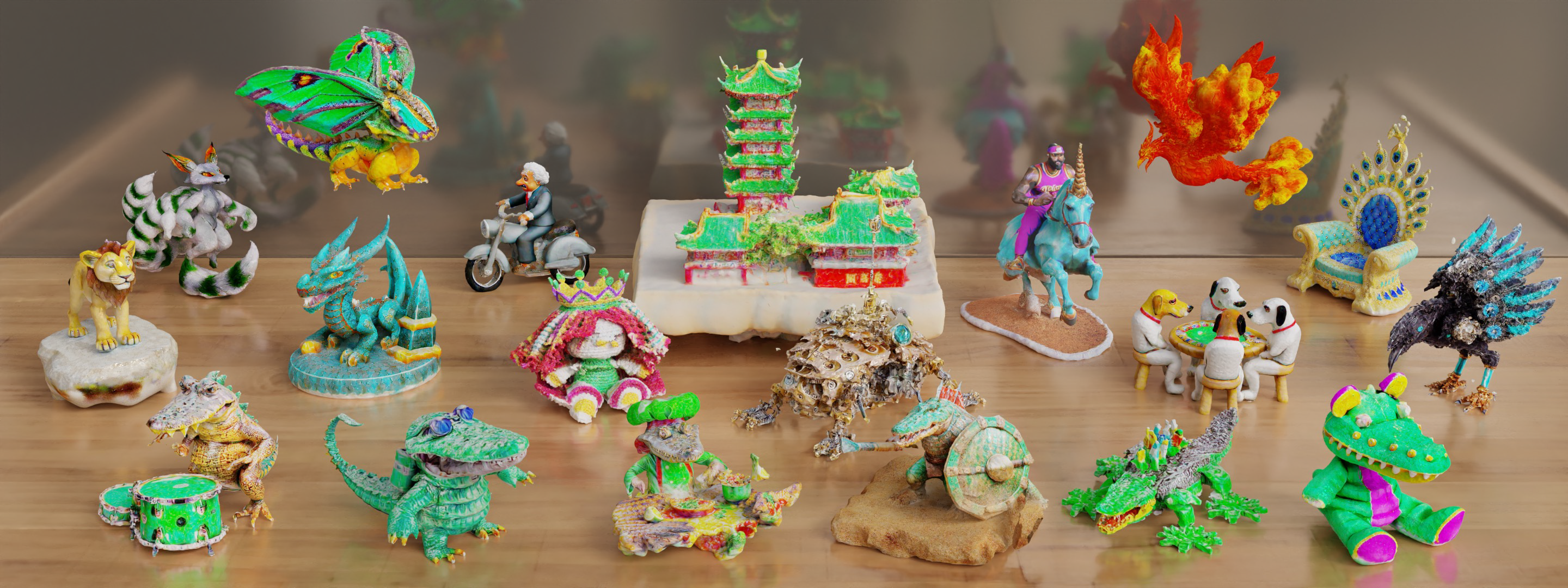}
    \caption{The examples of the 3D generated models with high quality and 4K details from RichDreamer \cite{qiu2023richdreamer}}
    \label{fig:richdreamerexample}
\end{figure*}

\section{Technical progress of the AI generated 3D content}

\subsection{Single 3D object generation}

In the realm of AI generated single 3D object, a variety of generative AI techniques are available. One approach directly uses point cloud diffusion models, exemplified by Point-E \cite{nichol2022pointe}. Another employs diffusion models trained in the wavelet domain, reconstructed using biorthogonal wavelet filters, as seen in EXIM \cite{liu2023exim}. The recent methods mostly involve generating multi-angle images using diffusion models, then transforming them into 3D models through NeRF \cite{mildenhall2020nerf} or 3DGS \cite{kerbl20233d}.

Methods for generating 3D models from multi-angle images generally fall into two main categories. The first involves iterative refinement to achieve detailed and coherent models in 3D space, usually requiring over an hour per model. This category often utilizes techniques like Score Distillation Sampling (SDS) to ensure overall model consistency. Research projects such as MVDream \cite{shi2023mvdream}, SweetDreamer \cite{li2023sweetdreamer}, and GaussianDreamer \cite{yi2023gaussiandreamer} exemplify this approach.

The second category involves neural networks generating multi-angle images in a single step, subsequently fitting these into 3D models. Examples include One-2-3-45++ \cite{liu2023onepp}, Instant3D \cite{li2023instant3d}, Wonder3D \cite{long2023wonder3d}, Consitent-1-to-3 \cite{ye2023consistent1to3}, and ZeroNVS \cite{sargent2023zeronvs}. These methods typically complete each model within a minute. However, they tend to produce lower-quality models, with resolutions around 128 to 256 cubed. Their 3D consistency originates from pre-trained 2D diffusion models, which, through fine-tuning, can generate coherent 2D images in 3D space. They often employ tri-planes techniques, using orthogonal planes in conjunction with neural networks to reconstruct 3D models. This method was introduced by EG3D in 2022 and further refined by Hexplane \cite{cao2023hexplane} and k-planes \cite{fridovichkeil2023kplanes} in early 2023.

It's important to note that most existing 3D AIGC models struggle with objects and scenes that include backgrounds. If provided with images containing backgrounds, objects must first be extracted before the 3D reconstruction process to avoid incorrect depth estimation. This limitation could be a drawback from 2D diffusion models, where the model's generalization capabilities make it challenging to generate a unified scene compared to a singular three-dimensional object. However, these issues are also being addressed by some papers. For instance the ZeroNVS \cite{sargent2023zeronvs} uses Denoising Diffusion Implicit Models (DDIM) \cite{song2022denoising} to tackle consistency problems. With this method employed, the scene around the object can be generated with more consistency.


Among the methods mentioned, One-2-3-45++ \cite{liu2023onepp} and One-2-3-45 \cite{liu2023one} represents generating one text to 3D model in 45 seconds. Their fine-tuned diffusion can output multiple angles of the same character on a single image. To achieve this, they employ a reference attention technique, which is one approach used in ControlNet \cite{zhang2023adding}, and one CLIP \cite{radford2021learning} image embedding as a Global Condition. With these modifications, they also fine-tune the Stable Diffusion2 \cite{rombach2021highresolution} on the Objaverse \cite{deitke2022objaverse} dataset. Subsequently, a 3D diffusion process follows the 2D multi-view diffusion to create a 3D voxel map. Their metrics in terms of F-score, CLIP-similarity, user preference, and time consumption outperform other methods. User preference is extensively utilized in this area, as the newly generated content lacks direct comparisons to datasets; most metrics are based on datasets or machine classifications derived from datasets. 

The Direct2.5 \cite{lu2023direct25} paper reports the shortest processing time among all discussed works, generating results in just ten seconds using the NVIDIA A100. The time calculation is divided into four steps, shown in Figure \ref{fig:direct25}. The first step involves generating four normal maps through multi-view normal diffusion, taking approximately 2.5 seconds. The second step combines the space carving algorithm \cite{kutulakos2000theory} with the marching cube algorithm to create a 3D mesh, iterating approximately 200 steps to optimize the loss function, which takes about 3.3 seconds. The third step, multi-view diffusion, employs a 4x super-resolution \cite{wang2021real} to produce improved images from 4 normal directions, taking around 2.5 seconds. The fourth step integrates textures with the mesh \cite{waechter2014let}. This multi-view diffusion utilizes self-attention to enhance consistency and employs DDIM to reduce noise. The authors also propose an idea of using holistic iteration for refining this model.

\begin{figure}
    \centering
    \includegraphics[width=0.485\textwidth]{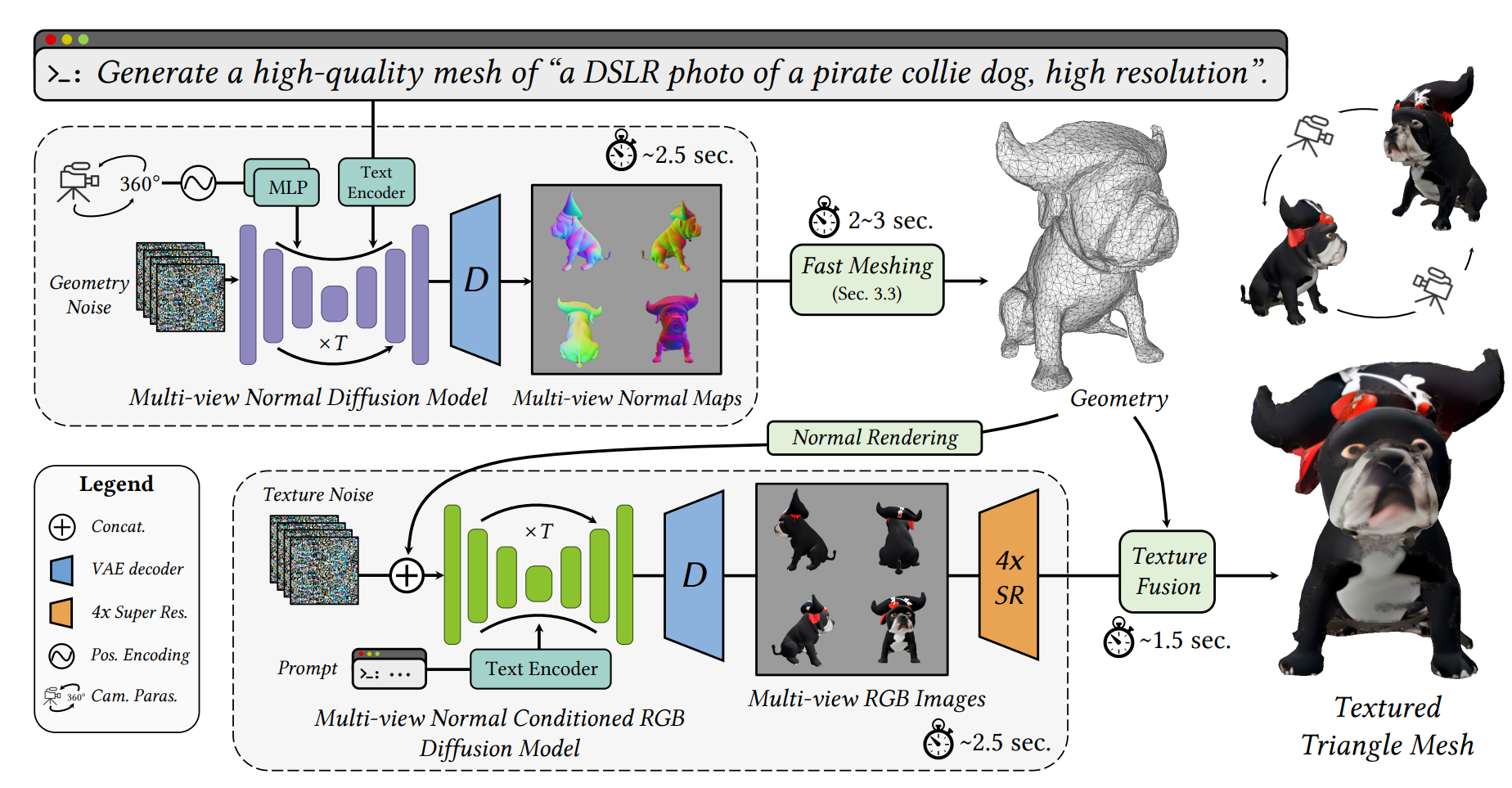}
    \caption{The structure graph of Direct2.5, especially specifies each step's time requirement, adding up to a total time of 10 seconds. Picture obtained from \cite{lu2023direct25}}
    \label{fig:direct25}
\end{figure}

Wonder3D \cite{long2023wonder3d} introduce one novel approach for generating 3D object from single image. It is built on concepts similar to previous paper. This paper combines multi-view and cross-domain attention with diffusion processes, addressing the challenges of maintaining consistency across multiple angles. Additionally, it features an updated loss function designed to enhance the detail of the generated output. The evaluation of this method utilizes Chamfer Distance and Volume Intersection over Union (IoU) as comparison standards, alongside commonly used image quality metrics such as PSNR, SSIM, and LPIPS. The model was trained on LVIS subset of the Objaverse dataset \cite{deitke2022objaverse} and extensively tested on the Google Scanned Object dataset \cite{downs2022google} with various 2D images of different styles. It only requires 2 to 3 minutes for each inference step.

The paper MVDream \cite{shi2023mvdream} presents an innovative approach using the Multi-view Diffusion Net, which incorporates attention mechanisms within the UNet architecture. The codebase utilizes the LDM model from Stable Diffusion. The multiview diffusion is initially trained on a 3D dataset to generate the first four views, serving as a pre-trained model for subsequent Score Distillation Sampling (SDS) processes. The multiview camera setup is encoded using a Multi-Layer Perceptron (MLP), allowing for the embedding into time or text dimensions. Through experimentation, embedding into the time dimension was found to be more robust. During the image generation phase, the SDS process is employed to fine-tune the results. In the final stage, DreamBooth \cite{ruiz2023dreambooth} is utilized to enhance consistency and generate additional views of the object, further augmenting the model's capability to produce detailed and consistent 3D representations from varied perspectives.

RichDreamer \cite{qiu2023richdreamer} stands out with the highest quality output among all the methods explored. This method leverages the LAION dataset \cite{schuhmann2022laion5b}, fully utilizing its capabilities for text-to-2D image generation. The LAION dataset's depth information significantly eases the training process and enhances 3D fusion. Building on the foundations set by MVDream \cite{shi2023mvdream} and Fantasia3D \cite{chen2023fantasia3d}, RichDreamer's primary innovation is the albedo diffusion model, along with a Normal-Depth diffusion model. The albedo diffusion model incorporates reflection parameters as a new loss parameter, enhancing the model's fidelity. Notably, the normal map, being more precise in 3D than the depth map, integrates seamlessly with normal maps from other viewpoints, thereby also improving the 3D awareness of the whole pipeline. Additionally, RichDreamer employs differentiable rasterization and Score Distillation Sampling (SDS) to optimize and refine 3D representations. In terms of experimental validation, this method achieved the highest CLIP scores and user study ratings. This paper also have a huge training process which requires a colossal 12,654 GPU hours on NVIDIA A100-80G GPUs, might be another reason why it work so well.

\begin{figure}
    \centering
    \includegraphics[width=0.485\textwidth]{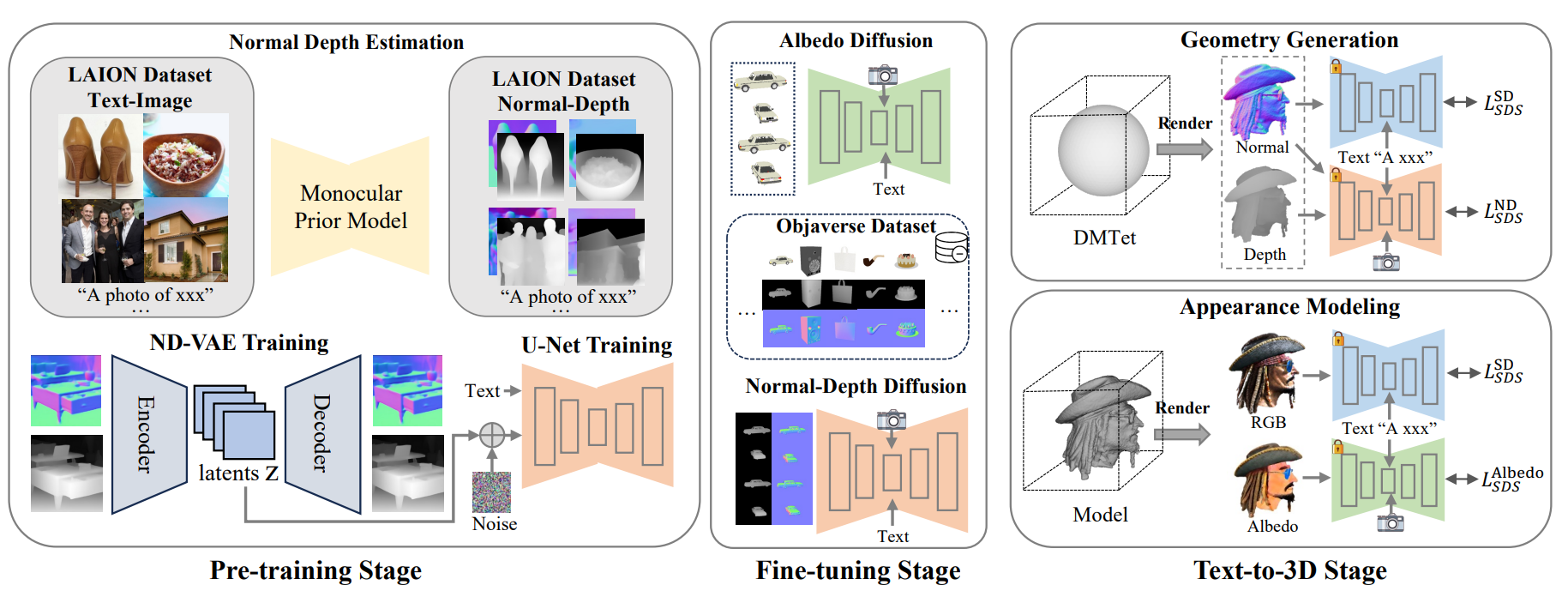}
    \caption{Structure for RichDreamer, picture obtained from \cite{qiu2023richdreamer}}
    \label{fig:enter-label}
\end{figure}

Another kind of methods like DN2N \cite{fang2023editing} allow for modifications on 3D models, altering appearances and expressions based on Diffusion, and could be integrated as steps in the processes mentioned above. A key feature of DN2N is its generalizable NeRF model architecture, which integrates cross-view regularization terms into the training process. This enhances 3D consistency in the edited novel views. The framework has been tested extensively across multiple datasets, demonstrating not only consistent 3D scene production but also diverse editing capabilities compared to other methods. Moreover, DN2N allows for a variety of editing types, including appearance editing, weather transformation, style transfer, and object changing, all while maintaining 3D consistency without needing a customized editing model for specific scenes. This approach significantly reduces the need for retraining for each new scene or editing type, thereby enhancing efficiency in terms of both runtime and model storage.

It's worth mentioning that in traditional 3D game and animation production, texture mapping was a critical consideration. However, with 2D Diffusion technology used into 3D objects generation, this issue has been largely mitigated, as multi-view image reconstructions inherently provide texture features for 3D assets. Nonetheless, fine-tuned texture modification remains essential for detailed application control. 

With the escalating demand for consistency in diffusion models, a variety of methods have been developed to address this problem. Techniques such as DreamBooth \cite{ruiz2023dreambooth}, LoRA \cite{hu2021lora}, self-attention \cite{vaswani2023attention}, CLIP embedding \cite{radford2021learning}, Zero123 \cite{liu2023zero1to3}, and ControlNet \cite{zhang2023adding} are notable examples. For example ControlNet which published in year 2023 has demonstrated its ability to control the original network with extra input parameters shown in Figure \ref{fig:controlnetpipe}. It has been proven especially useful for controlling pictures generated from the diffusion network. With its simplicity and extensibility, it might be able to extend to many other parameters like camera pose or 3D sketching. With these methods continuously improve consistency, they are not only solving the problem in 2D AI generation but 3D AI generation as well. For instance, Stable Video Diffusion (SVD) is claimed to be capable of generating multi-angle views of an object from a singular image and text prompt, thereby indirectly achieving consistency in 3D with multi-angle 2D videos. These evolving methodologies are gradually reshaping the landscape of content generation in both 2D and 3D domains.

\begin{figure}
    \centering
    \includegraphics[width=0.485\textwidth]{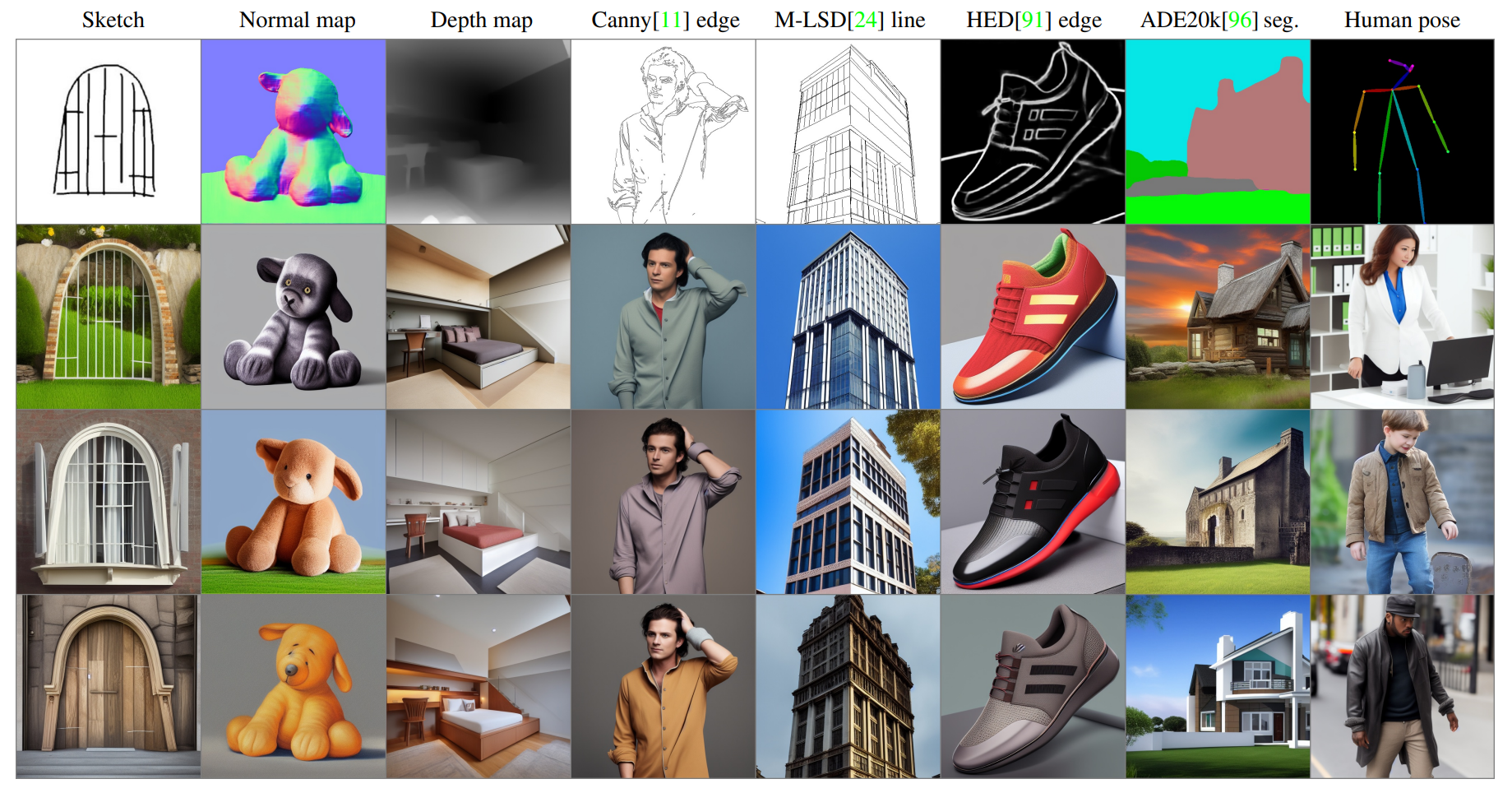}
    \caption{Example of different control modes from ControlNet, picture obtained from \cite{zhang2023adding}}
    \label{fig:controlnetexample}
\end{figure}

\begin{figure}
    \centering
    \includegraphics[width=0.485\textwidth]{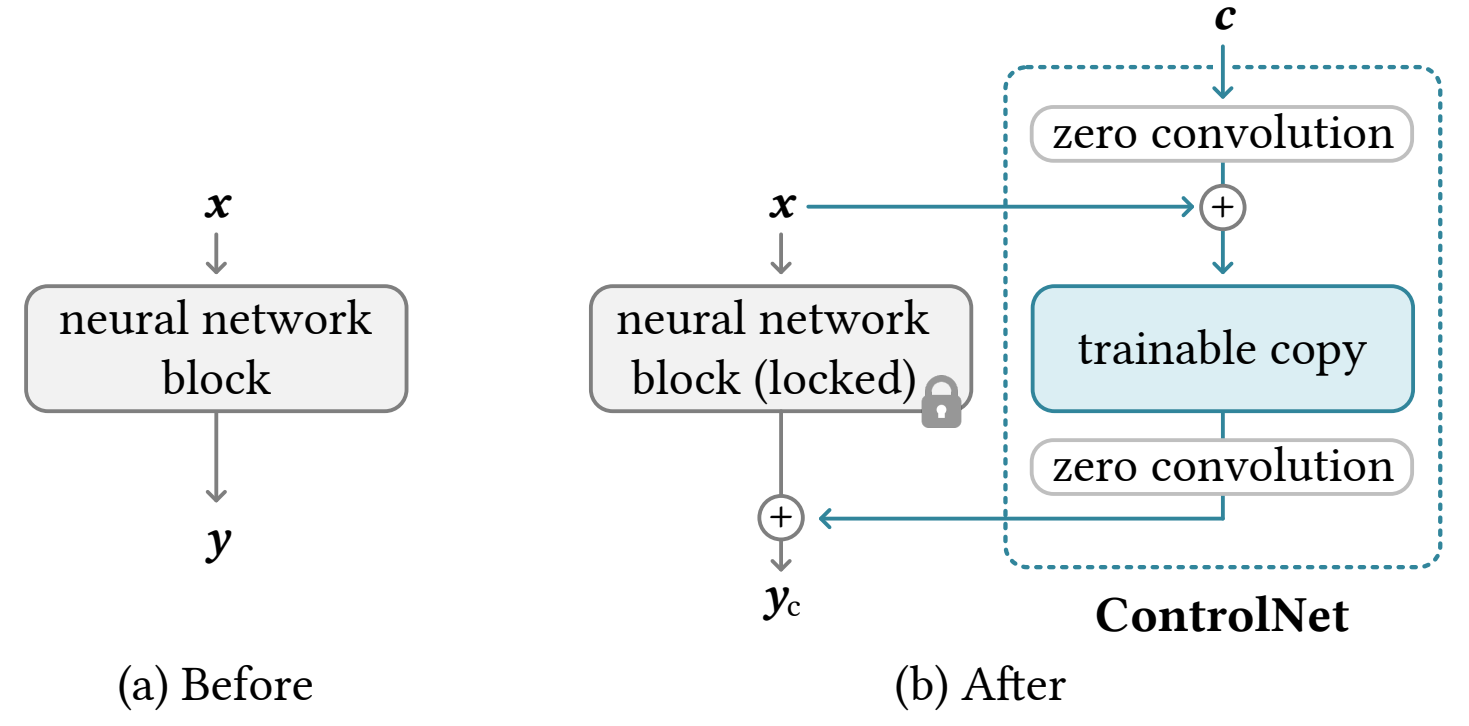}
    \caption{Structure of ControlNet, picture obtained from \cite{zhang2023adding}}
    \label{fig:controlnetpipe}
\end{figure}

Papers like Zero123 \cite{liu2023zero1to3} and the later updated Stable Zero123 \cite{stablezero123} have shown promising results in generating novel views of a single object using only one image input along with a new rotation information. The rotation input has two degrees of freedom.  An important aspect of this model is the incorporation of rotation in the CLIP embedding \cite{radford2021learning}. With adequate training, these models have proven to be quite effective, allowing 3D modeling and change of view.

\subsection{3D human model generation}

In the area of 3D human reconstruction using AI, the categorization is similar to that of model reconstruction. There are methods that generate high-precision human models through iteration, as well as those that produce relatively rougher human models in a single pass, often referred to as Avatars. To date, I haven't encountered any research papers in 3D human generation that excellently handle both human models and motions simultaneously. However, there are noteworthy achievements in each area separately, with human motion generation being a topic for a subsequent major chapter.

3D human generation is inherently more complex due to the involvement of facial expressions and clothing textures, and typically, unlike static 3D object models, these need to be animate. Techniques commonly employed include the SMPL \cite{loper2015smpl} and its successor, SMPL-X \cite{pavlakos2019smplx}, widely used for human modeling and training prior. Which decrease the difficulty in human body modeling. SMPL comprises an external 3D mesh, a movable skeleton rig, and blended poses and movements. Its benefits are manifold; it offers more details than standard 3D object files and serves as a neutral human body dataset, eliminating numerous training steps that would otherwise start from gaussian noise. This model supports Python format exports like NPZ and PKL and is compatible with platforms like Unity and Blender. It's also supported by open-source viewing methods, including aitviewer \cite{Kaufmann_Vechev_aitviewer_2022}.

There are many iterative human generation methods, such as DreamWaltz \cite{huang2023dreamwaltz}, HumanNorm \cite{huang2023humannorm}, TADA \cite{liao2023tada}, TeCH \cite{huang2023tech}, DreamAvatar \cite{cao2023dreamavatar}, etc. For instance, DreamWaltz uses SMPL \cite{loper2015smpl} as a human precursor, generating human skeletal structures and images from multiple angles. Those data are input into ControlNet and optimized using Stable Diffusion \cite{rombach2021highresolution}. Using SDS iteration for further modeling optimization, this method can create a human model on a NVIDIA RTX 3090 with about an hour, while subsequent animation generation is faster, at 3 seconds per frame. Although DreamWaltz lacks specific comparative metrics, its’s user study is best among all compared avatar creation models. Its iterative nature means further refinement is possible with more training time.

HumanNorm \cite{huang2023humannorm} trains separate depth-adapted and norm-adapted diffusion models, as shown in Figure \ref{fig:humannormexp}. They use DMTET \cite{shen2021deep} for 3D mesh reconstruction. It introduces Progressive Geometry Generation, initially focusing on low-frequency components and gradually lifting the mask onto high-frequency components for improved results. The SDS process is based on both normal and depth map. It also optimizes SDS with Coarse-to-fine texture generation, and overcome the oversaturated result.  For each generation it require a NVIDIA RTX 3090 for two hours, it produces 1024x1024 images and videos with clear facial features compared to other methods. Its unique approach gave it factual accuracy and expressive range. Its resulting human model is much superior than other methods and can be seen in its comparison photos and user study. Some of its examples is shown in Figure \ref{fig:humannormexp}.

\begin{figure*}
    \centering
    \includegraphics[width=1\textwidth]{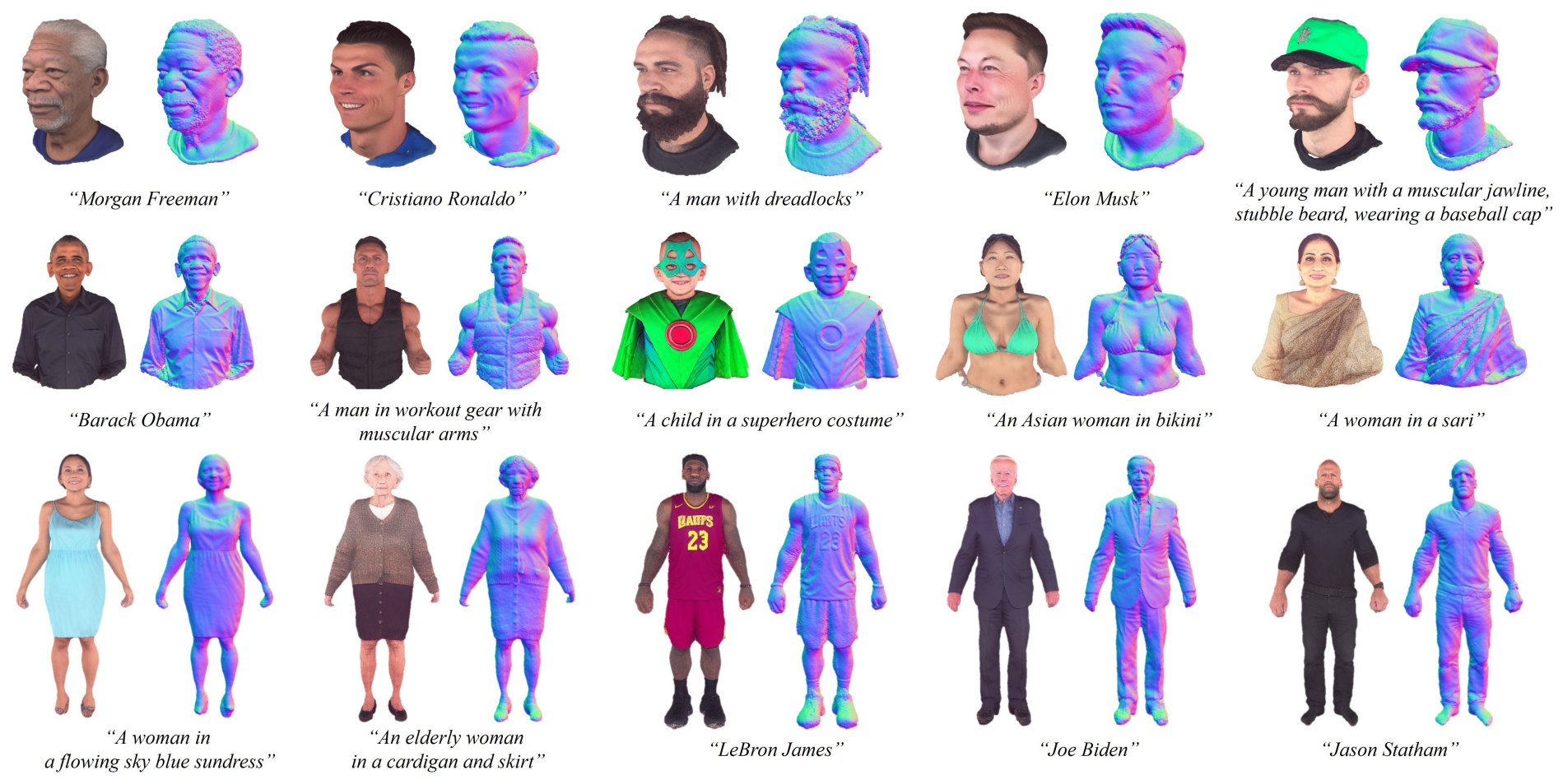}
    \caption{The generated examples from HumanNorm, its progressive generation from low frequency to high frequency results in high fidelity 3D human models, with realistic facial details, picture obtained from \cite{huang2023humannorm} }
    \label{fig:humannormexp}
\end{figure*}

\begin{figure}
    \centering
    \includegraphics[width=0.485\textwidth]{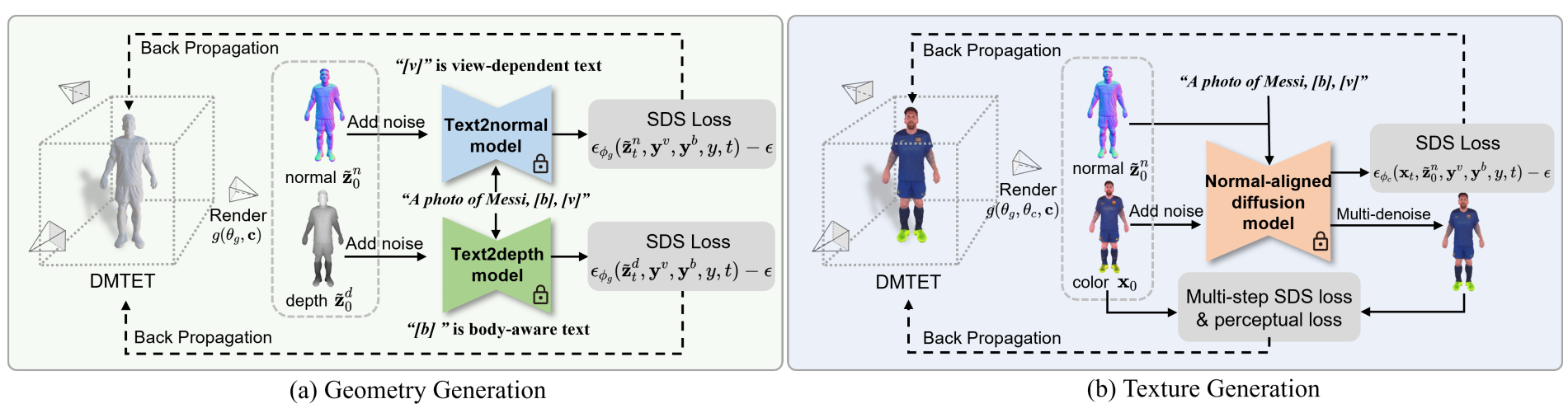}
    \caption{Structure of HumanNorm,  picture obtained from \cite{huang2023humannorm}}
    \label{fig:humannormexp}
\end{figure}

Non-iterative human generation methods like GTA \cite{zhang2023globalcorrelated}, Chupa \cite{kim2023chupa}, AvatarCraft \cite{jiang2023avatarcraft}, DreamHuman \cite{kolotouros2023dreamhuman}, and One-shot Implicit Animatable Avatars with Model-based Priors \cite{huang2023oneshot} is also noteworthy. GTA is particularly distinguished for its rapid texture modeling capabilities. This method facilitates the generation of a 3D human model from a single image, employing a tri-plane approach to ascertain the planes perpendicular to the input image. In order to learn the perpendicular planes with regard to the input image plane. The paper introduce a learnable embedding z, which is refined through self-attention encoding and subsequently utilized as a query in multi-head cross-attention mechanisms. As a non-iterative technique, GTA's outcomes are commendable for their quality. GTA's comparative standards include Chamfer, P2S, Normals, assessing surface-to-point distances. Also includes and PSNR used for accessing image reconstruction quality.

Another method Chupa \cite{kim2023chupa}, which can output 3D human models based on text input, but is also capable of generating random human models in the absence of text. This approach is based on the SMPL-X \cite{pavlakos2019smplx} model and Stable Diffusion as foundation. The process, shown in Figure \ref{fig:chupapipe}, begins with creating a frontal image using Stable Diffusion based on text input, or noise if no text is provided. This image, along with the frontal view of the initial SMPL-X human model, is fed into a latent diffusion model to generate images of both the front and back of the model. Subsequently, the Neural Deferred Shading method is used to reconstruct the 3D human model, complete with rudimentary hair, clothing, and facial expressions. Individual modifications of the head and overall model are then made using SDEdit \cite{meng2022sdedit}. Finally, photos of the model from multiple angles (totaling 36 in experiments) are adjusted using SDEdit and reconstructed using Neural Density Field (NDF). Compared to the three hours required by AvatarCLIP \cite{hong2022avatarclip}, this method produces a model in just three minutes on an RTX 3090. Despite the involved multi-angle adjustments, Chupa is still classified as a non-iterative method due to its relatively few modifications, unlike other methods that may involve tens of thousands of iterations.

DreamHuman \cite{kolotouros2023dreamhuman} employs NeRF as the 3D representation for its generative model, functioning as an end-to-end neural network that facilitates iterative training. The method initially generates a base for pose and body shape using imGHUM \cite{alldieck2021imghum}. It then utilizes semantic zooming technology to focus on visually important body areas, such as the face and hands, thereby enhancing overall detail and improving the quality of the generated model. As the training involves an end-to-end network, this paper requires a comprehensive set of loss function parameters, including density loss from imGHUM \cite{alldieck2021imghum}, predicted normal loss and orientation loss from Ref-NeRF \cite{verbin2021refnerf}, loss on the proposal weights from mip-NeRF360 \cite{barron2022mipnerf}, and foreground mask loss. Compared to its predecessors, this paper achieves more precise control over body movements and enhances local details through semantic zooming.

\subsection{3D Scene generation}

Research papers have made considerable progress in the generation of 3D scenes, although not as extensively as previous sections. There 
are a variety of approaches to tackle the AI generated Scene. PERF \cite{wang2023perf} is capable of generating 3D scenes from a single panoramic image, supplementing shadow areas using Diffusion models. ZeroNVS \cite{sargent2023zeronvs} goes a step further by reconstructing both objects and environments in 3D based on a single image. While its environmental reconstruction is somewhat less detailed, the algorithm demonstrates an understanding of environmental contexts. EXIM \cite{liu2023exim}, on the other hand, is capable of generating scenes such as living rooms and kitchens, partly due to its use of the ShapeNet dataset, which includes a wide array of corresponding furniture models. LucidDreamer \cite{chung2023luciddreamer} is one paper that also utilize the 2D diffusion model for extended the area of input image, its generation is only 2.5D but the result still works very well with the 3DGS, shown in Figure \ref{fig:luciddreamer}.

\begin{figure}
    \centering
    \includegraphics[width=0.485\textwidth]{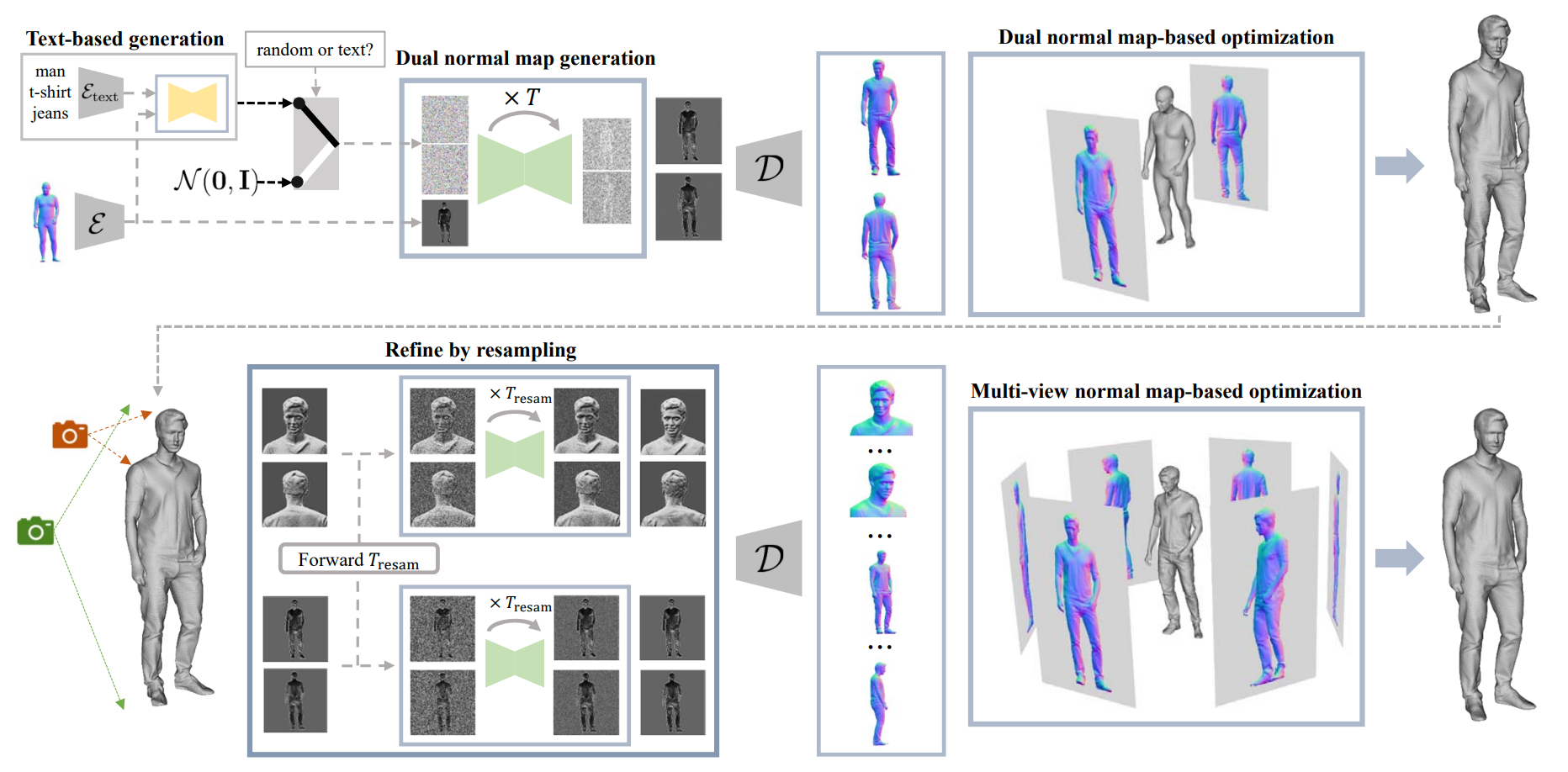}
    \caption{Structure of Chupa, utilizing diffusion for detailed refinement, the diffusion generated normal maps exhibit high fidelity, picture obtained from  \cite{kim2023chupa}}
    \label{fig:chupapipe}
\end{figure}

SceneDreamer \cite{chen2023scenedreamer} is the first to generate unbounded 3D scenes from a collection of 2D images as shown in Figure \ref{fig:scenedreamer}, without relying on 3D annotations. It uses an unconditional generative model that synthesizes 3D landscapes from random noises. The core elements include a Bird’s-Eye-View (BEV) scene representation, which efficiently represents surface elevation and scene semantics, a semantic-aware neural hash grid for parameterizing varied latent features, and a style-based volumetric renderer that produces photo-realistic images. This innovative method addresses significant challenges in AI generated 3D scene, such as efficient representation of unbounded scenes and alignment of content across various scales and orientations. SceneDreamer stands out for its ability to generate large-scale, diverse 3D worlds, marking a first in the domain of unbounded 3D scene generation from 2D images.

\begin{figure*}
    \centering
    \includegraphics[width=1\textwidth]{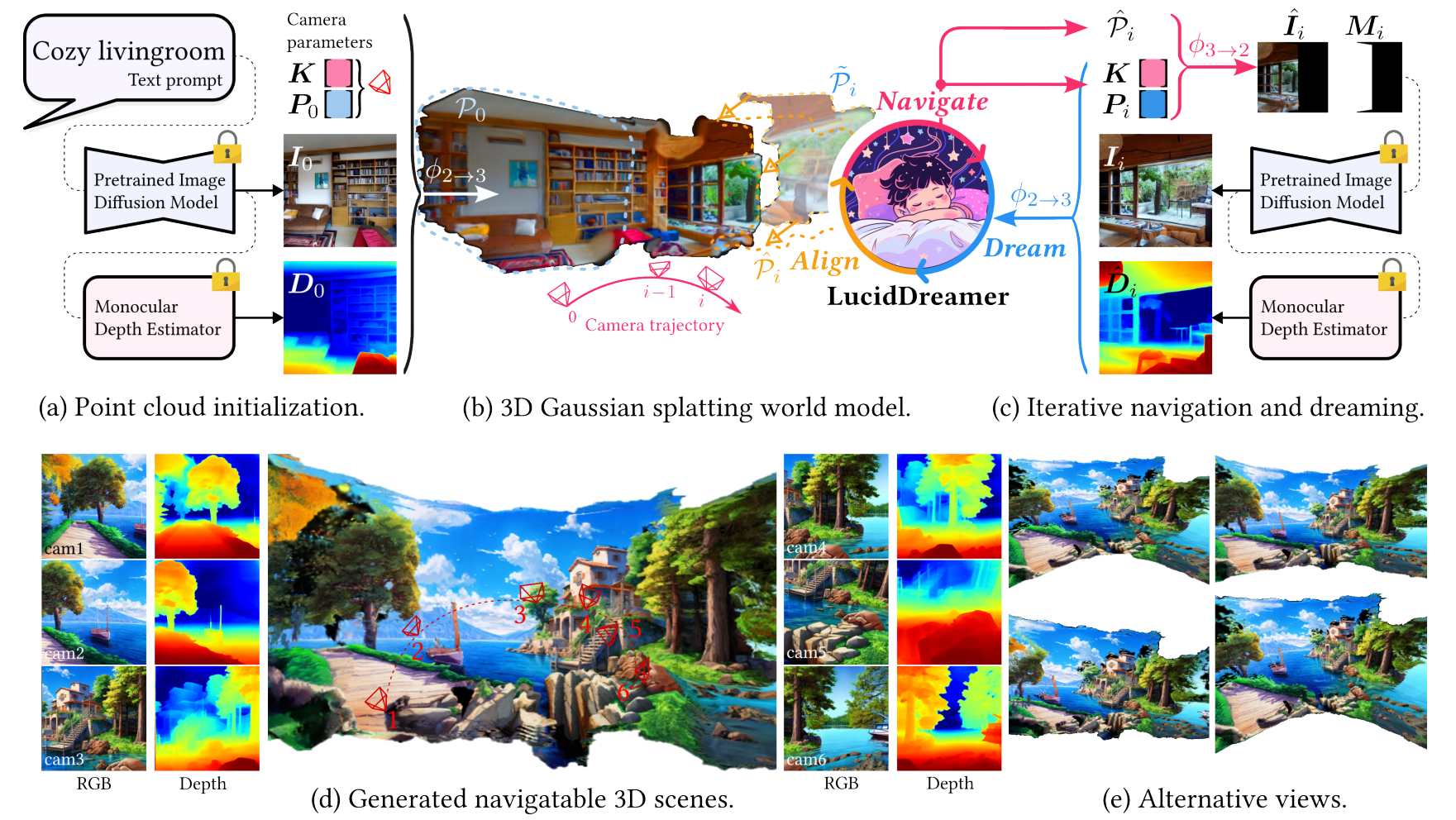}
    \caption{The result and methods from LicidDreamer shows its capability in generate 2.5D scene with 3D Gaussian Splatting, picture obtained from \cite{chung2023luciddreamer}}
    \label{fig:luciddreamer}
\end{figure*}

PERF \cite{wang2023perf} is an innovative method that enables the creation of 3D scenes from a single panoramic image. It particularly addressing challenges in synthesizing 360-degree views with large occlusions. Unlike previous works that focused on limited fields of view and fewer occlusions, PERF introduces a collaborative RGBD inpainting method and a progressive inpainting-and-erasing strategy. These techniques allow for the completion of both RGB images and depth maps, generating novel appearances and 3D shapes that are not visible in the input panorama. The approach begins with predicting a panoramic depth map from a single panorama, followed by reconstructing visible 3D regions. In addition, PERF employs a Stable Diffusion model for RGB inpainting and a monocular depth estimator for depth completion. This combination of techniques ensures consistency across different views and effectively addresses geometry conflicts and unseen area of the given panoramic image. Extensive experiments have demonstrated PERF's superiority over existing methods, making it suitable for a range of applications like panorama-to-3D conversion, text-to-3D rendering, and 3D scene stylization. Maybe due to the lack of training data, its text to 3D scene will result in jagged edge in many places.

ZeroNVS \cite{sargent2023zeronvs}, Zero-Shot 360-Degree View Synthesis from a Single Real Image, an advancement over previous method Zero1-2-3. While the previous method is mainly focus on 3D object generation. This one can generate 3D scene with a virtual rotating camera. The model is trained on a diverse dataset mix, including CO3D, RealEstate10K, and ACID, which encompass a variety of scenes from object-centric to indoor and outdoor settings. This method addresses the limited diversity issue observed in Score Distillation Sampling (SDS), particularly in the generation of scene backgrounds. Standard SDS tends to produce monotonous backgrounds, especially in long-range novel view synthesis. To counter this, they introduce SDS anchoring which involves initially sampling several "anchor" novel views using the Denoising Diffusion Implicit Model (DDIM). These views, with their diverse contents, serve as a more varied base for SDS, ensuring diversity while maintaining 3D consistency in the synthesized views. 

\begin{figure}
    \centering
    \includegraphics[width=0.485\textwidth]{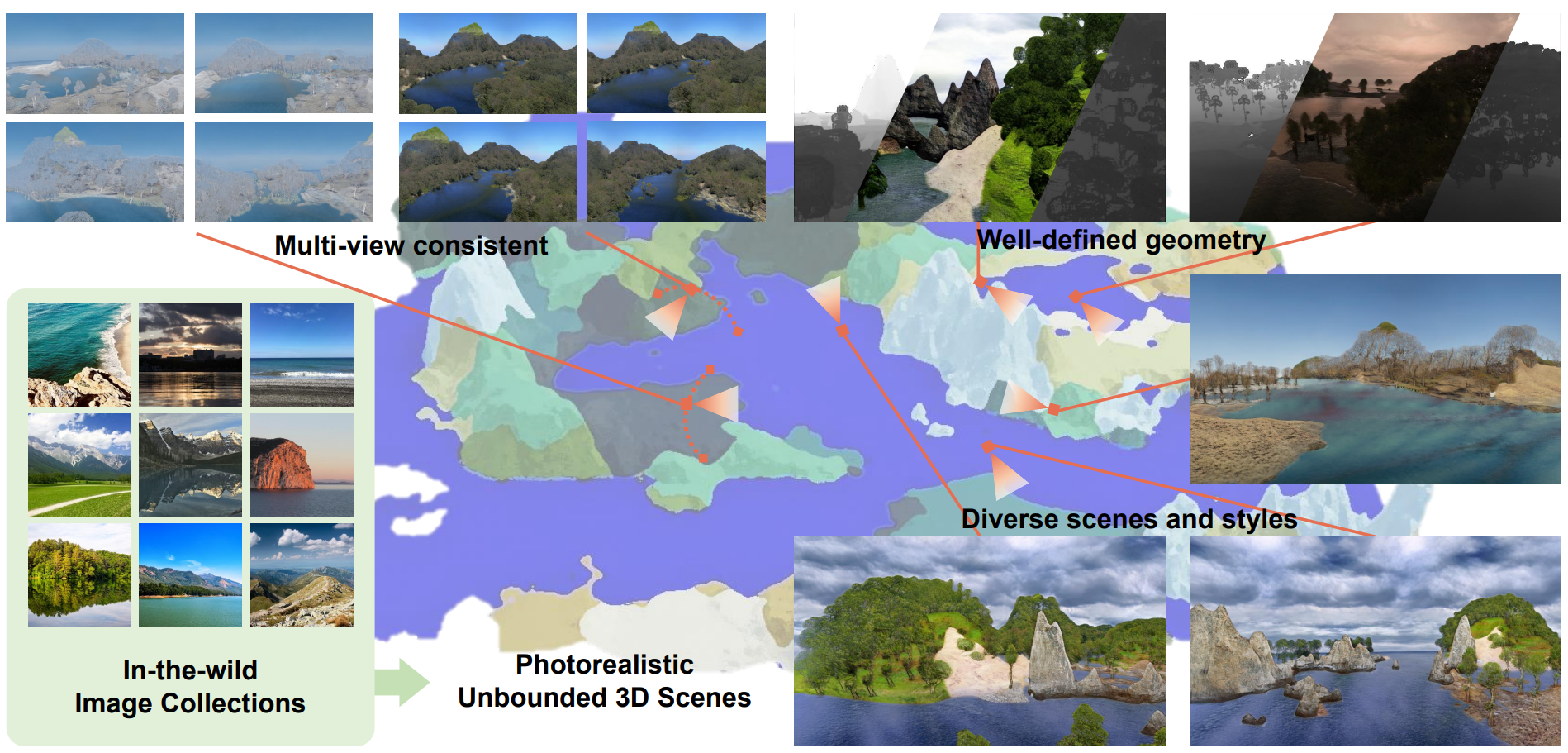}
    \caption{Example of SceneDreamer's showing it can embed the input image into a variety of location during the generation of a 3D world, picture obtained from \cite{chen2023scenedreamer}}
    \label{fig:scenedreamer}
\end{figure}

There are techniques that exclusively utilize 3D datasets for training, such as EXIM \cite{liu2023exim}, which incorporates biorthogonal wavelet filters in its methodology for generating 3D shapes from textual descriptions. This method employs a hybrid approach combining explicit and implicit representations. In the explicit stage, it controls the 3D shape's topological structure, leveraging biorthogonal wavelet filters for shape reconstruction, and permits text aware local modifications. The implicit stage refines these shapes further and introduces color and texture. This pipeline ensure a harmonious consistency between them. A notable strength of EXIM is its capacity to produce high-fidelity shapes. However, its dependence on 3D datasets for training could limit its generalization capabilities, potentially limiting its application in broader scenes or environments.

\subsection{Human motion synthesis}

There are several notable papers on generating human motion, such as DIMOS \cite{zhao2023synthesizing}, Story2Motion \cite{qing2023storytomotion}, HumanTOMATO \cite{lu2023humantomato}, and Learned Motion Matching \cite{holden2020learned}. DIMOS is a highly effective system that can guide motions based on marked points or perform movements such as sitting, lying down, and standing up. It is notable for its impressive result, as it not only facilitates human movement but also enables a certain degree of interaction with objects. The model divides its training into two parts: movement and object interaction. For the interaction aspect, it leverages the capabilities from the preceding paper, COINS, to enable meaningful interactions. The movement component is compared with GAMMA, scoring impressively with a 0.99 for foot-ground contact and avoiding non-contact areas, demonstrating its effectiveness. The network primarily utilizes GRU and ResNet, resembling a custom-built network architecture. Before use, DIMOS requires a pre-constructed scene setup, including chairs, beds, sofas, and obstacles. The combination of two prior studies enables the system to integrate human movement with motions, and it can even be controlled through text. Despite its high scores, there are still minor issues with object penetration in its practical applications.

The paper Story2Motion \cite{qing2023storytomotion} achieves the generation of motion from extensive text solely using a Transformer-based Large Language Model (LLM) and a human motion database, as shown in Figure \ref{fig:story2motion}. It discerns the phases and sequences of motions from the text, filling gaps with walking motions and then seamlessly connecting them to form a complete motion sequence. The motion scheduler part uses a pre-trained LLM to extract semantic information from the input story, creating a series of (text, position, duration) pairs. Then they utilize a motion database to retrieve and blend the motion. The motion retrieved should conform both semantic and trajectory constraints. It utilizes progressive generation, attention masks, and positional encoding for the motion segment blending. This represents one of the few methods that rely mainly on Transformers to generate all motions. The effectiveness of this approach is notable, as it demonstrates a superior understanding of text, enabling a broader range of motions compared to other previous works. Scene and semantic information about the scene are required to be provided in advance.

The papers HumanTOMATO \cite{lu2023humantomato} and T2M-GPT \cite{zhang2023t2mgpt} both employ a structure that combines Transformer with Vector Quantized Variational AutoEncoder (VQ-VAE) for motion synthesis. The advantage of these methods lie in its capability to fully accomplish motion synthesis through the neural network. Originally, VQ-VAE was utilized to train images by encoding a model corresponding to the images. Now, it is adapted for training to align textual parameters with motions, enabling its decoder to interpret the output of the Transformer as specific movements. As it has the ability to compress the motion into the VQ-VAE, it does not require a separate motion dataset for motion synthesis. The T2M-GPT also addresses challenges such as code collapse in VQ-VAE training, using strategies like Exponential Moving Average (EMA) and Code Reset, and alleviating discrepancies between training and inference in GPT with simple sequence corruption during training. 

Both "Plan, Posture and Go" \cite{liu2023plan}, and InsActor \cite{ren2023insactor} utilize the diffusion model for the motion synthesis. "Plan, Posture and Go"'s pipeline works like this. It first leverages the GPT3.5 to refine text input into detailed descriptions of poses. This method then utilize the diversity of diffusion models to generate optimal candidate postures. Finally, the go-diffuser model, based on diffusion, will add rotation and translation between the selected postures. Outputting a full pose sequence.It exhibits improved performance metrics, such as better R value, FID, and MultiModal Distance, compared to its predecessors. Which indicates its continuity is highly optimized.

Advancing further, the generation of human-object interaction animations poses even greater challenges. One of the more successful methods is IMoS \cite{ghosh2023imos}, which can produce full-body 3D animations with objects based on input motions, objects, and textual instructions. However, it is not flawless, particularly in perfecting contact points. Another notable paper worth mentioning is InterDiff \cite{xu2023interdiff}, which can determine the contact points between a person and an object based on their initial positions and then allows the object to follow the movement of the human through displacement. The trained network significantly reduces instances of clipping, where the human model and object intersect unrealistically. Despite this, the final interactions still have some areas that lack realism. CHORE \cite{xie2023chore}, on the other hand, predicts the positional relationship between humans and objects frame by frame based on images, which results in the loss of temporal 3D consistency.

\section{Disscussion}

\subsection{Advancement of 3D rendering}

In the field of 3D Scene Description, the use of neural networks for scene storage has become a prominent trend. For example, NeRF \cite{mildenhall2020nerf} can be used for render entire scenes, requiring only a few images and camera parameters as initial input. To obtain the output from the network, just use a new camera parameter, and the network directly outputs an image. Compared to traditional methods, NeRF bypasses manual scene construction, ray tracing, and the effects of lighting and reflection. It can be considered as a direct mapping from input camera parameters to output images. However, NeRF has its own technical pain points, primarily due to that scene representation is implicit inside the neural nework. Previously it has to retraining the neural network to change the scene. Some new studies have shown  adding a modifying neural network at the end or extracting and modifying object Mesh information can be used for modifying the NeRF network \cite{liu2021editing,yuan2022nerfediting,chen2023neuraleditor}.

\begin{figure*}
    \centering
    \includegraphics[width=1\textwidth]{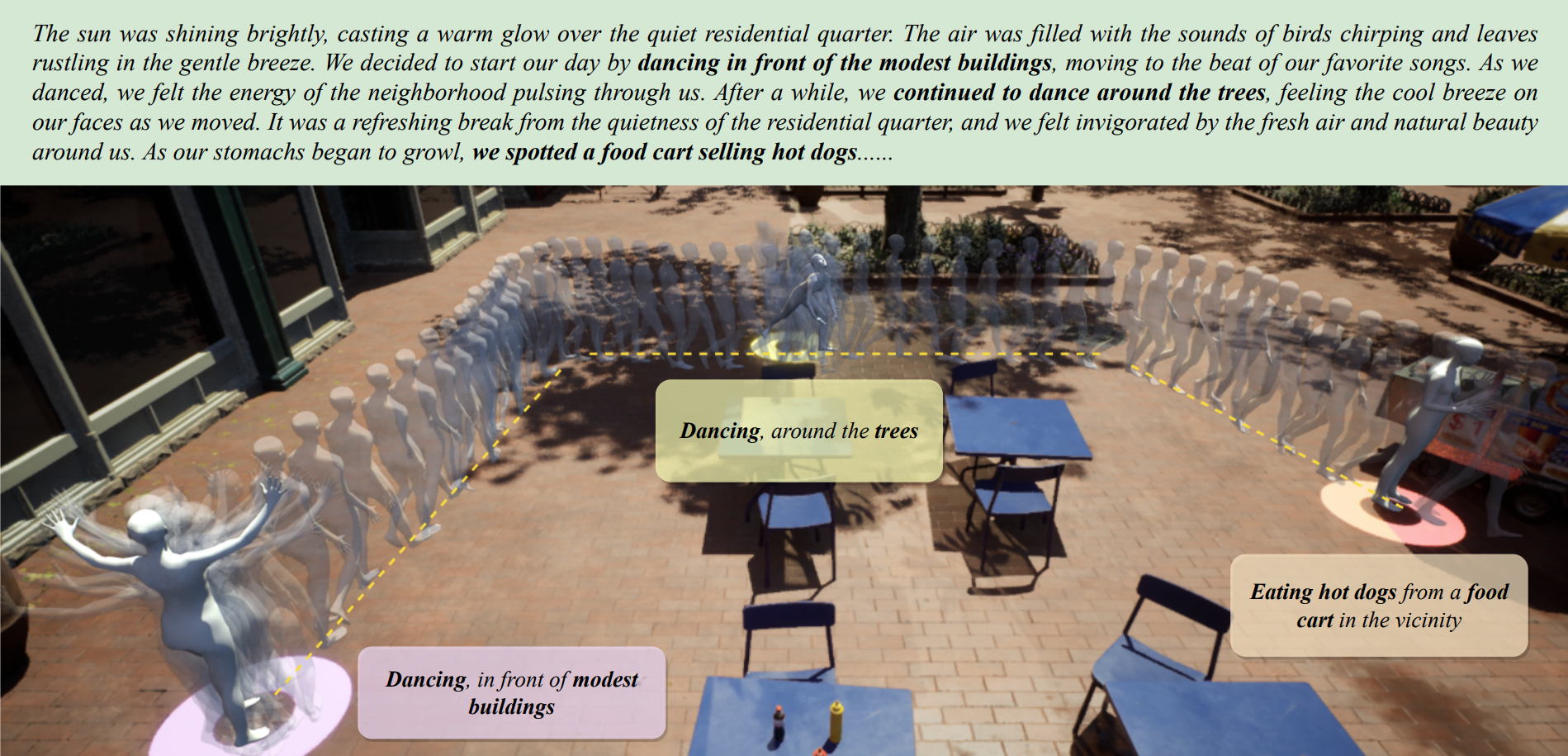}
    \caption{Story2Motion demonstrates its capability of generating a sequence of human motion with regard to a long sequence of text input. Picture obtained from  \cite{qing2023storytomotion}}
    \label{fig:story2motion}
\end{figure*}

Gaussian Splatting describes scenes using 3D Gaussian points, where each point's position is fixed, but its descriptive range can vary in size, containing color and opacity information. With training, Gaussian points can be added, split, or deleted. In terms of post-processing, adding, deleting, or scaling in 3DGS is relatively easier; theoretically, the color of 3D Gaussian points can also be directly modified. However, because each point's descriptive 3D range is not certain, although it provides compression space for neural networks, it introduces many uncertainties in modification, which are problems that need to be solved. Another well-known method is Neural Deferred Shading (NDS), which is faster than other methods in generating images based on viewpoints. However, it can only reconstruct 3D for mesh objects, limiting its application to human bodies or objects.

Many exciting improvements also have been pushed forward for those two methods. Google released SMERF \cite{duckworth2023smerf}. Based on NeRF, it not only has higher clarity than 3DGS but also supports streaming transmission for real-time rendering with about 144+ FPS. Another method 4K4D \cite{xu20234k4d} can also do 4k rendering with 80+ FPS. FSGS can also help rendering scene using 3DGS at real time. Those new methods has greatly improve the usability of the emerging AI rendering technique. Methods like SpacetimeGaussians \cite{li2023spacetime} and 4DGS \cite{wu20234d}, not only enhance the clarity of 3DGS but also render the entire scene animatable, facilitating the viewing experience of 3D movies. Recent approaches such as Deblurring 3DGS \cite{lee2024deblurring} have been developed to address the clarity issues associated with 3D Gaussian Splatting (3DGS).

These advancements have paved the way for neural networks to bridge the gap between image and 3D, as the content can go either way easily. This opens up more possibilities for the future of Graphics, as well as in the area of AR and VR. However, in terms of generation and artistry, without human involvement, it is still not possible to achieve sufficient clarity and precision of expression. If AI generation can further advance in terms of human control, it could compensate for the excessive randomness and lack of continuity inherent in AI-generated content.

\subsection{Solving the fidelity and accuracy}

Generation of 3D Objects or Scenes often relies on the capability to generate 2D images. This involves using Diffusion or GAN image generation methods. In these two fields, newly released papers like Stable Diffusion XL \cite{podell2023sdxl} and GigaGAN \cite{kang2023scaling} represent significant advancements. Although currently, the usage rate of Diffusion methods far surpass those of GANs in the 3D generation area. But as seen in GigaGAN, the GAN methods are better suited for high-frequency details, with their main advantages being the generation speed and precision. While the diffusion is especially good at handling low-frequency signals. A future combination of these two directions is also possible. However, maintaining sufficient consistency across multiple images remains the biggest challenge in 3D generation.

Another promising progress for 3D generation progress is the increasing of 3D datasets. With sufficient datasets and efficient training methods, training on 3D datasets could potentially achieve much better zero-shot effects, allowing neural networks to generate new content that is not included in their training datasets. The main technical challenge in 3D scene generation is the lack of datasets compared to 3D models, leading to lower precision and factual errors in environment details. 

Human Models and motions are often stored using SMPL-X \cite{pavlakos2019smplx}, which serves not only as a storage model but also as a process method for estimating 3D human models from single image input to software output. Other papers typically use it as a precursor, modifying human details such as hair, clothing, expressions, and motions based on it. With this kind of preliminary model, coupled with the precision of AI-generated character images, AI-generated 3D humans can now achieve greater clarity and accuracy compared to traditional 3D models. Having a skeletal model of the character also makes animating the figure relatively simpler.

In motion synthesis, understanding scenes and pairing text with motions often involve foundational models like Transformer and VQ-VAE. Transformers are highly versatile for text input and output and can handle sequential information in motions. VQ-VAE excels at compressing information into a discrete codebook. As a result, many methods interpret the codebook through Transformers and then train VQ-VAE to align the codebook with motion data. VQ-VAE's Encoder-Decoder also has compression capabilities to record motion data. Models using only Transformers rely on motion data from datasets. Technical challenges in this area focus on issues like foot sliding and penetration, which traditional methods can handle. For instance, Learned Motion Matching \cite{holden2020learned}, which combines traditional methods with neural networks, effectively addresses these issues. However, using neural networks alone can lead to boundary issues, as they approximate boundaries using loss function, making precise resolution difficult. Another challenge is semantically describing content in scenes and moving characters accordingly. For sufficiently good effects, methods like Story2Motion use preset scene descriptions. Existing models like Open-NeRF \cite{zhang2023opennerf} can accurately describe objects in scenes, but integrating this into the motion generation process remains a challenge.

This field still lacks a precise metrics, with many papers relying on blind user tests for comparison. Some papers like One-2-3-45++ \cite{liu2023onepp} and Wonder3D \cite{long2023wonder3d}  use 2D image metrics for comparison. 
I think From a user perspective, scoring metrics worth considering include: the model's generalization capability, 3D model precision, 3D texture detail, the proportion of factual errors noticeable to the human eye, and understanding of input text or images.

In the evolving field of  AI generated 3D content, there still lack a precise metrics. Currently, many studies relying on blind user tests for comparison. Papers like One-2-3-45++ \cite{liu2023onepp} and Wonder3D \cite{long2023wonder3d} adapt 2D image metrics for assessment. While these methods offer some insights, there could be a benefit in developing metrics that are more tailored to the unique aspects of 3D modeling. From a user's perspective, potential metrics to consider could include the model's ability to generalize across different scenarios, the accuracy of the 3D models, the quality of the 3D textures, and the degree to which any errors are perceptible to an observer. Another important factor is the model's capability to interpret input texts or images effectively.

\subsection{Potential applications}

The applications of these models are quite varied. It holds particular promise in the gaming industry.  They offer an opportunity to streamline the workflow for 3D artists, potentially simplifying the transition from concept to model and easing the process of model refinement. This might allow the artists to focus on enhancing 3D models using refined 2D images. Additionally, the replication and generalization capabilities of these generative models might enrich game environments, offering a variety of elements like trees, flowers, and buildings, thereby adding to the visual diversity.

In sectors like education and advertising, 3D generation holds significant potential due to the strong spatial understanding and memory humans have for 3D objects. Educational content or advertisements presented in 3D might be more engaging and memorable for audiences, similar to how physical experiments in a physics class can enhance understanding. Furthermore, with the increasing integration of VR and AR technologies, there's a growing need for software that allows multi-angle observation of objects, opening up new avenues for these models to make an impact.

\bibliographystyle{unsrt}  
\bibliography{references} 

\begin{thebibliography}{10}

\bibitem{mildenhall2020nerf}
Ben Mildenhall, Pratul~P. Srinivasan, Matthew Tancik, Jonathan~T. Barron, Ravi
  Ramamoorthi, and Ren Ng.
\newblock Nerf: Representing scenes as neural radiance fields for view
  synthesis, 2020.

\bibitem{kerbl20233d}
Bernhard Kerbl, Georgios Kopanas, Thomas Leimkühler, and George Drettakis.
\newblock 3d gaussian splatting for real-time radiance field rendering, 2023.

\bibitem{li2023generative}
Chenghao Li, Chaoning Zhang, Atish Waghwase, Lik-Hang Lee, Francois Rameau,
  Yang Yang, Sung-Ho Bae, and Choong~Seon Hong.
\newblock Generative ai meets 3d: A survey on text-to-3d in aigc era, 2023.

\bibitem{shi2023mvdream}
Yichun Shi, Peng Wang, Jianglong Ye, Mai Long, Kejie Li, and Xiao Yang.
\newblock Mvdream: Multi-view diffusion for 3d generation, 2023.

\bibitem{qiu2023richdreamer}
Lingteng Qiu, Guanying Chen, Xiaodong Gu, Qi~Zuo, Mutian Xu, Yushuang Wu,
  Weihao Yuan, Zilong Dong, Liefeng Bo, and Xiaoguang Han.
\newblock Richdreamer: A generalizable normal-depth diffusion model for detail
  richness in text-to-3d, 2023.

\bibitem{lu2023direct25}
Yuanxun Lu, Jingyang Zhang, Shiwei Li, Tian Fang, David McKinnon, Yanghai Tsin,
  Long Quan, Xun Cao, and Yao Yao.
\newblock Direct2.5: Diverse text-to-3d generation via multi-view 2.5d
  diffusion, 2023.

\bibitem{deitke2023objaversexl}
Matt Deitke, Ruoshi Liu, Matthew Wallingford, Huong Ngo, Oscar Michel, Aditya
  Kusupati, Alan Fan, Christian Laforte, Vikram Voleti, Samir~Yitzhak Gadre,
  Eli VanderBilt, Aniruddha Kembhavi, Carl Vondrick, Georgia Gkioxari, Kiana
  Ehsani, Ludwig Schmidt, and Ali Farhadi.
\newblock Objaverse-xl: A universe of 10m+ 3d objects, 2023.

\bibitem{wang2023diffusiondb}
Zijie~J. Wang, Evan Montoya, David Munechika, Haoyang Yang, Benjamin Hoover,
  and Duen~Horng Chau.
\newblock Diffusiondb: A large-scale prompt gallery dataset for text-to-image
  generative models, 2023.

\bibitem{loper2015smpl}
Matthew Loper, Naureen Mahmood, Javier Romero, Gerard Pons-Moll, and Michael~J.
  Black.
\newblock {SMPL}: A skinned multi-person linear model.
\newblock {\em ACM Trans. Graphics (Proc. SIGGRAPH Asia)}, 34(6):248:1--248:16,
  October 2015.

\bibitem{pavlakos2019smplx}
Georgios Pavlakos, Vasileios Choutas, Nima Ghorbani, Timo Bolkart, Ahmed A.~A.
  Osman, Dimitrios Tzionas, and Michael~J. Black.
\newblock Expressive body capture: {3D} hands, face, and body from a single
  image.
\newblock In {\em Proceedings IEEE Conf. on Computer Vision and Pattern
  Recognition (CVPR)}, pages 10975--10985, 2019.

\bibitem{qing2023storytomotion}
Zhongfei Qing, Zhongang Cai, Zhitao Yang, and Lei Yang.
\newblock Story-to-motion: Synthesizing infinite and controllable character
  animation from long text, 2023.

\bibitem{zhang2023globalcorrelated}
Zechuan Zhang, Li~Sun, Zongxin Yang, Ling Chen, and Yi~Yang.
\newblock Global-correlated 3d-decoupling transformer for clothed avatar
  reconstruction, 2023.

\bibitem{nichol2022pointe}
Alex Nichol, Heewoo Jun, Prafulla Dhariwal, Pamela Mishkin, and Mark Chen.
\newblock Point-e: A system for generating 3d point clouds from complex
  prompts, 2022.

\bibitem{liu2023exim}
Zhengzhe Liu, Jingyu Hu, Ka-Hei Hui, Xiaojuan Qi, Daniel Cohen-Or, and Chi-Wing
  Fu.
\newblock Exim: A hybrid explicit-implicit representation for text-guided 3d
  shape generation, 2023.

\bibitem{li2023sweetdreamer}
Weiyu Li, Rui Chen, Xuelin Chen, and Ping Tan.
\newblock Sweetdreamer: Aligning geometric priors in 2d diffusion for
  consistent text-to-3d, 2023.

\bibitem{yi2023gaussiandreamer}
Taoran Yi, Jiemin Fang, Junjie Wang, Guanjun Wu, Lingxi Xie, Xiaopeng Zhang,
  Wenyu Liu, Qi~Tian, and Xinggang Wang.
\newblock Gaussiandreamer: Fast generation from text to 3d gaussians by
  bridging 2d and 3d diffusion models, 2023.

\bibitem{liu2023onepp}
Minghua Liu, Ruoxi Shi, Linghao Chen, Zhuoyang Zhang, Chao Xu, Xinyue Wei,
  Hansheng Chen, Chong Zeng, Jiayuan Gu, and Hao Su.
\newblock One-2-3-45++: Fast single image to 3d objects with consistent
  multi-view generation and 3d diffusion.
\newblock {\em arXiv preprint arXiv:2311.07885}, 2023.

\bibitem{li2023instant3d}
Jiahao Li, Hao Tan, Kai Zhang, Zexiang Xu, Fujun Luan, Yinghao Xu, Yicong Hong,
  Kalyan Sunkavalli, Greg Shakhnarovich, and Sai Bi.
\newblock Instant3d: Fast text-to-3d with sparse-view generation and large
  reconstruction model, 2023.

\bibitem{long2023wonder3d}
Xiaoxiao Long, Yuan-Chen Guo, Cheng Lin, Yuan Liu, Zhiyang Dou, Lingjie Liu,
  Yuexin Ma, Song-Hai Zhang, Marc Habermann, Christian Theobalt, and Wenping
  Wang.
\newblock Wonder3d: Single image to 3d using cross-domain diffusion, 2023.

\bibitem{ye2023consistent1to3}
Jianglong Ye, Peng Wang, Kejie Li, Yichun Shi, and Heng Wang.
\newblock Consistent-1-to-3: Consistent image to 3d view synthesis via
  geometry-aware diffusion models, 2023.

\bibitem{sargent2023zeronvs}
Kyle Sargent, Zizhang Li, Tanmay Shah, Charles Herrmann, Hong-Xing Yu, Yunzhi
  Zhang, Eric~Ryan Chan, Dmitry Lagun, Li~Fei-Fei, Deqing Sun, and Jiajun Wu.
\newblock Zeronvs: Zero-shot 360-degree view synthesis from a single real
  image, 2023.

\bibitem{cao2023hexplane}
Ang Cao and Justin Johnson.
\newblock Hexplane: A fast representation for dynamic scenes, 2023.

\bibitem{fridovichkeil2023kplanes}
Sara Fridovich-Keil, Giacomo Meanti, Frederik Warburg, Benjamin Recht, and
  Angjoo Kanazawa.
\newblock K-planes: Explicit radiance fields in space, time, and appearance,
  2023.

\bibitem{song2022denoising}
Jiaming Song, Chenlin Meng, and Stefano Ermon.
\newblock Denoising diffusion implicit models, 2022.

\bibitem{liu2023one}
Minghua Liu, Chao Xu, Haian Jin, Linghao Chen, Zexiang Xu, Hao Su, et~al.
\newblock One-2-3-45: Any single image to 3d mesh in 45 seconds without
  per-shape optimization.
\newblock {\em arXiv preprint arXiv:2306.16928}, 2023.

\bibitem{zhang2023adding}
Lvmin Zhang, Anyi Rao, and Maneesh Agrawala.
\newblock Adding conditional control to text-to-image diffusion models.
\newblock In {\em Proceedings of the IEEE/CVF International Conference on
  Computer Vision}, pages 3836--3847, 2023.

\bibitem{radford2021learning}
Alec Radford, Jong~Wook Kim, Chris Hallacy, Aditya Ramesh, Gabriel Goh,
  Sandhini Agarwal, Girish Sastry, Amanda Askell, Pamela Mishkin, Jack Clark,
  Gretchen Krueger, and Ilya Sutskever.
\newblock Learning transferable visual models from natural language
  supervision, 2021.

\bibitem{rombach2021highresolution}
Robin Rombach, Andreas Blattmann, Dominik Lorenz, Patrick Esser, and Björn
  Ommer.
\newblock High-resolution image synthesis with latent diffusion models, 2021.

\bibitem{deitke2022objaverse}
Matt Deitke, Dustin Schwenk, Jordi Salvador, Luca Weihs, Oscar Michel, Eli
  VanderBilt, Ludwig Schmidt, Kiana Ehsani, Aniruddha Kembhavi, and Ali
  Farhadi.
\newblock Objaverse: A universe of annotated 3d objects, 2022.

\bibitem{kutulakos2000theory}
Kiriakos~N Kutulakos and Steven~M Seitz.
\newblock A theory of shape by space carving.
\newblock {\em International journal of computer vision}, 38:199--218, 2000.

\bibitem{wang2021real}
Xintao Wang, Liangbin Xie, Chao Dong, and Ying Shan.
\newblock Real-esrgan: Training real-world blind super-resolution with pure
  synthetic data.
\newblock In {\em Proceedings of the IEEE/CVF international conference on
  computer vision}, pages 1905--1914, 2021.

\bibitem{waechter2014let}
Michael Waechter, Nils Moehrle, and Michael Goesele.
\newblock Let there be color! large-scale texturing of 3d reconstructions.
\newblock In {\em Computer Vision--ECCV 2014: 13th European Conference, Zurich,
  Switzerland, September 6-12, 2014, Proceedings, Part V 13}, pages 836--850.
  Springer, 2014.

\bibitem{downs2022google}
Laura Downs, Anthony Francis, Nate Koenig, Brandon Kinman, Ryan Hickman, Krista
  Reymann, Thomas~B. McHugh, and Vincent Vanhoucke.
\newblock Google scanned objects: A high-quality dataset of 3d scanned
  household items, 2022.

\bibitem{ruiz2023dreambooth}
Nataniel Ruiz, Yuanzhen Li, Varun Jampani, Yael Pritch, Michael Rubinstein, and
  Kfir Aberman.
\newblock Dreambooth: Fine tuning text-to-image diffusion models for
  subject-driven generation, 2023.

\bibitem{schuhmann2022laion5b}
Christoph Schuhmann, Romain Beaumont, Richard Vencu, Cade Gordon, Ross
  Wightman, Mehdi Cherti, Theo Coombes, Aarush Katta, Clayton Mullis, Mitchell
  Wortsman, Patrick Schramowski, Srivatsa Kundurthy, Katherine Crowson, Ludwig
  Schmidt, Robert Kaczmarczyk, and Jenia Jitsev.
\newblock Laion-5b: An open large-scale dataset for training next generation
  image-text models, 2022.

\bibitem{chen2023fantasia3d}
Rui Chen, Yongwei Chen, Ningxin Jiao, and Kui Jia.
\newblock Fantasia3d: Disentangling geometry and appearance for high-quality
  text-to-3d content creation, 2023.

\bibitem{fang2023editing}
Shuangkang Fang, Yufeng Wang, Yi~Yang, Yi-Hsuan Tsai, Wenrui Ding, Shuchang
  Zhou, and Ming-Hsuan Yang.
\newblock Editing 3d scenes via text prompts without retraining, 2023.

\bibitem{hu2021lora}
Edward~J. Hu, Yelong Shen, Phillip Wallis, Zeyuan Allen-Zhu, Yuanzhi Li, Shean
  Wang, Lu~Wang, and Weizhu Chen.
\newblock Lora: Low-rank adaptation of large language models, 2021.

\bibitem{vaswani2023attention}
Ashish Vaswani, Noam Shazeer, Niki Parmar, Jakob Uszkoreit, Llion Jones,
  Aidan~N. Gomez, Lukasz Kaiser, and Illia Polosukhin.
\newblock Attention is all you need, 2023.

\bibitem{liu2023zero1to3}
Ruoshi Liu, Rundi Wu, Basile~Van Hoorick, Pavel Tokmakov, Sergey Zakharov, and
  Carl Vondrick.
\newblock Zero-1-to-3: Zero-shot one image to 3d object, 2023.

\bibitem{stablezero123}
{Stability AI}.
\newblock {Stable Zero123: Quality 3D Object Generation from Single Images}.
\newblock \url{https://stability.ai/news/stable-zero123-3d-generation}, 2023.
\newblock Online; accessed 13 December 2023.

\bibitem{Kaufmann_Vechev_aitviewer_2022}
Manuel Kaufmann, Velko Vechev, and Dario Mylonopoulos.
\newblock {aitviewer}, 7 2022.

\bibitem{huang2023dreamwaltz}
Yukun Huang, Jianan Wang, Ailing Zeng, He~Cao, Xianbiao Qi, Yukai Shi,
  Zheng-Jun Zha, and Lei Zhang.
\newblock Dreamwaltz: Make a scene with complex 3d animatable avatars, 2023.

\bibitem{huang2023humannorm}
Xin Huang, Ruizhi Shao, Qi~Zhang, Hongwen Zhang, Ying Feng, Yebin Liu, and Qing
  Wang.
\newblock Humannorm: Learning normal diffusion model for high-quality and
  realistic 3d human generation, 2023.

\bibitem{liao2023tada}
Tingting Liao, Hongwei Yi, Yuliang Xiu, Jiaxaing Tang, Yangyi Huang, Justus
  Thies, and Michael~J. Black.
\newblock Tada! text to animatable digital avatars, 2023.

\bibitem{huang2023tech}
Yangyi Huang, Hongwei Yi, Yuliang Xiu, Tingting Liao, Jiaxiang Tang, Deng Cai,
  and Justus Thies.
\newblock Tech: Text-guided reconstruction of lifelike clothed humans, 2023.

\bibitem{cao2023dreamavatar}
Yukang Cao, Yan-Pei Cao, Kai Han, Ying Shan, and Kwan-Yee~K. Wong.
\newblock Dreamavatar: Text-and-shape guided 3d human avatar generation via
  diffusion models, 2023.

\bibitem{shen2021deep}
Tianchang Shen, Jun Gao, Kangxue Yin, Ming-Yu Liu, and Sanja Fidler.
\newblock Deep marching tetrahedra: a hybrid representation for high-resolution
  3d shape synthesis, 2021.

\bibitem{kim2023chupa}
Byungjun Kim, Patrick Kwon, Kwangho Lee, Myunggi Lee, Sookwan Han, Daesik Kim,
  and Hanbyul Joo.
\newblock Chupa: Carving 3d clothed humans from skinned shape priors using 2d
  diffusion probabilistic models, 2023.

\bibitem{jiang2023avatarcraft}
Ruixiang Jiang, Can Wang, Jingbo Zhang, Menglei Chai, Mingming He, Dongdong
  Chen, and Jing Liao.
\newblock Avatarcraft: Transforming text into neural human avatars with
  parameterized shape and pose control, 2023.

\bibitem{kolotouros2023dreamhuman}
Nikos Kolotouros, Thiemo Alldieck, Andrei Zanfir, Eduard~Gabriel Bazavan, Mihai
  Fieraru, and Cristian Sminchisescu.
\newblock Dreamhuman: Animatable 3d avatars from text, 2023.

\bibitem{huang2023oneshot}
Yangyi Huang, Hongwei Yi, Weiyang Liu, Haofan Wang, Boxi Wu, Wenxiao Wang,
  Binbin Lin, Debing Zhang, and Deng Cai.
\newblock One-shot implicit animatable avatars with model-based priors, 2023.

\bibitem{meng2022sdedit}
Chenlin Meng, Yutong He, Yang Song, Jiaming Song, Jiajun Wu, Jun-Yan Zhu, and
  Stefano Ermon.
\newblock Sdedit: Guided image synthesis and editing with stochastic
  differential equations, 2022.

\bibitem{hong2022avatarclip}
Fangzhou Hong, Mingyuan Zhang, Liang Pan, Zhongang Cai, Lei Yang, and Ziwei
  Liu.
\newblock Avatarclip: Zero-shot text-driven generation and animation of 3d
  avatars, 2022.

\bibitem{alldieck2021imghum}
Thiemo Alldieck, Hongyi Xu, and Cristian Sminchisescu.
\newblock imghum: Implicit generative models of 3d human shape and articulated
  pose, 2021.

\bibitem{verbin2021refnerf}
Dor Verbin, Peter Hedman, Ben Mildenhall, Todd Zickler, Jonathan~T. Barron, and
  Pratul~P. Srinivasan.
\newblock Ref-nerf: Structured view-dependent appearance for neural radiance
  fields, 2021.

\bibitem{barron2022mipnerf}
Jonathan~T. Barron, Ben Mildenhall, Dor Verbin, Pratul~P. Srinivasan, and Peter
  Hedman.
\newblock Mip-nerf 360: Unbounded anti-aliased neural radiance fields, 2022.

\bibitem{wang2023perf}
Guangcong Wang, Peng Wang, Zhaoxi Chen, Wenping Wang, Chen~Change Loy, and
  Ziwei Liu.
\newblock Perf: Panoramic neural radiance field from a single panorama, 2023.

\bibitem{chung2023luciddreamer}
Jaeyoung Chung, Suyoung Lee, Hyeongjin Nam, Jaerin Lee, and Kyoung~Mu Lee.
\newblock Luciddreamer: Domain-free generation of 3d gaussian splatting scenes,
  2023.

\bibitem{chen2023scenedreamer}
Zhaoxi Chen, Guangcong Wang, and Ziwei Liu.
\newblock Scenedreamer: Unbounded 3d scene generation from 2d image
  collections.
\newblock {\em IEEE Transactions on Pattern Analysis and Machine Intelligence},
  45(12):15562–15576, December 2023.

\bibitem{zhao2023synthesizing}
Kaifeng Zhao, Yan Zhang, Shaofei Wang, Thabo Beeler, and Siyu Tang.
\newblock Synthesizing diverse human motions in 3d indoor scenes, 2023.

\bibitem{lu2023humantomato}
Shunlin Lu, Ling-Hao Chen, Ailing Zeng, Jing Lin, Ruimao Zhang, Lei Zhang, and
  Heung-Yeung Shum.
\newblock Humantomato: Text-aligned whole-body motion generation, 2023.

\bibitem{holden2020learned}
Daniel Holden, Oussama Kanoun, Maksym Perepichka, and Tiberiu Popa.
\newblock Learned motion matching.
\newblock {\em ACM Transactions on Graphics (TOG)}, 39(4):53--1, 2020.

\bibitem{zhang2023t2mgpt}
Jianrong Zhang, Yangsong Zhang, Xiaodong Cun, Shaoli Huang, Yong Zhang, Hongwei
  Zhao, Hongtao Lu, and Xi~Shen.
\newblock T2m-gpt: Generating human motion from textual descriptions with
  discrete representations, 2023.

\bibitem{liu2023plan}
Jinpeng Liu, Wenxun Dai, Chunyu Wang, Yiji Cheng, Yansong Tang, and Xin Tong.
\newblock Plan, posture and go: Towards open-world text-to-motion generation,
  2023.

\bibitem{ren2023insactor}
Jiawei Ren, Mingyuan Zhang, Cunjun Yu, Xiao Ma, Liang Pan, and Ziwei Liu.
\newblock Insactor: Instruction-driven physics-based characters, 2023.

\bibitem{ghosh2023imos}
Anindita Ghosh, Rishabh Dabral, Vladislav Golyanik, Christian Theobalt, and
  Philipp Slusallek.
\newblock Imos: Intent-driven full-body motion synthesis for human-object
  interactions, 2023.

\bibitem{xu2023interdiff}
Sirui Xu, Zhengyuan Li, Yu-Xiong Wang, and Liang-Yan Gui.
\newblock Interdiff: Generating 3d human-object interactions with
  physics-informed diffusion, 2023.

\bibitem{xie2023chore}
Xianghui Xie, Bharat~Lal Bhatnagar, and Gerard Pons-Moll.
\newblock Chore: Contact, human and object reconstruction from a single rgb
  image, 2023.

\bibitem{liu2021editing}
Steven Liu, Xiuming Zhang, Zhoutong Zhang, Richard Zhang, Jun-Yan Zhu, and
  Bryan Russell.
\newblock Editing conditional radiance fields, 2021.

\bibitem{yuan2022nerfediting}
Yu-Jie Yuan, Yang-Tian Sun, Yu-Kun Lai, Yuewen Ma, Rongfei Jia, and Lin Gao.
\newblock Nerf-editing: Geometry editing of neural radiance fields, 2022.

\bibitem{chen2023neuraleditor}
Jun-Kun Chen, Jipeng Lyu, and Yu-Xiong Wang.
\newblock Neuraleditor: Editing neural radiance fields via manipulating point
  clouds, 2023.

\bibitem{duckworth2023smerf}
Daniel Duckworth, Peter Hedman, Christian Reiser, Peter Zhizhin, Jean-François
  Thibert, Mario Lučić, Richard Szeliski, and Jonathan~T. Barron.
\newblock Smerf: Streamable memory efficient radiance fields for real-time
  large-scene exploration, 2023.

\bibitem{xu20234k4d}
Zhen Xu, Sida Peng, Haotong Lin, Guangzhao He, Jiaming Sun, Yujun Shen, Hujun
  Bao, and Xiaowei Zhou.
\newblock 4k4d: Real-time 4d view synthesis at 4k resolution, 2023.

\bibitem{li2023spacetime}
Zhan Li, Zhang Chen, Zhong Li, and Yi~Xu.
\newblock Spacetime gaussian feature splatting for real-time dynamic view
  synthesis, 2023.

\bibitem{wu20234d}
Guanjun Wu, Taoran Yi, Jiemin Fang, Lingxi Xie, Xiaopeng Zhang, Wei Wei, Wenyu
  Liu, Qi~Tian, and Xinggang Wang.
\newblock 4d gaussian splatting for real-time dynamic scene rendering, 2023.

\bibitem{lee2024deblurring}
Byeonghyeon Lee, Howoong Lee, Xiangyu Sun, Usman Ali, and Eunbyung Park.
\newblock Deblurring 3d gaussian splatting, 2024.

\bibitem{podell2023sdxl}
Dustin Podell, Zion English, Kyle Lacey, Andreas Blattmann, Tim Dockhorn, Jonas
  Müller, Joe Penna, and Robin Rombach.
\newblock Sdxl: Improving latent diffusion models for high-resolution image
  synthesis, 2023.

\bibitem{kang2023scaling}
Minguk Kang, Jun-Yan Zhu, Richard Zhang, Jaesik Park, Eli Shechtman, Sylvain
  Paris, and Taesung Park.
\newblock Scaling up gans for text-to-image synthesis, 2023.

\bibitem{zhang2023opennerf}
Hao Zhang, Fang Li, and Narendra Ahuja.
\newblock Open-nerf: Towards open vocabulary nerf decomposition, 2023.

\end{thebibliography}

\end{twocolumn}

\end{document}